\documentclass[journal]{IEEEtran}     % Use this line for a4
\ifCLASSINFOpdf
\else
\fi
\usepackage{graphics} % for pdf, bitmapped graphics files
\usepackage{epsfig} % for postscript graphics files
\usepackage{mathptmx} % assumes new font selection scheme installed
\usepackage{times} % assumes new font selection scheme installed
\usepackage{amsmath} % assumes amsmath package installed
\usepackage{amssymb}  % assumes amsmath package installed
\usepackage{color}
\usepackage{fancyvrb}
\usepackage{alltt}
\usepackage{url}
\usepackage{algorithm}
\usepackage{algpseudocode}
\usepackage{multirow}
\usepackage{subfig}
\usepackage{epstopdf}

\graphicspath{{}}

\renewcommand\tabcolsep{3pt}

\begin{document}
	
\title{High-level Reasoning and Low-level Learning for Grasping: A Probabilistic Logic Pipeline}
\author{Laura Antanas, Plinio Moreno, Marion Neumann, Rui Pimentel de Figueiredo, Kristian Kersting, Jos\'{e} Santos-Victor, Luc De Raedt}

%The paper headers

\maketitle

%%%%%%%%%%%%%%%%%%%%%%%%%%%%%%%%%%%%%%%%%%%%%%%%%%%%%%%%%%%%%%%%%%%%%%%%%%%%%%%%
\begin{abstract}
While grasps must satisfy the grasping stability criteria, good grasps depend on the specific manipulation scenario: the object, its properties and functionalities, as well as the task and grasp constraints. In this paper, we consider such information for robot grasping by \emph{leveraging manifolds} and \emph{symbolic object parts}.  Specifically, we introduce a new \emph{probabilistic logic} module to first \emph{semantically reason} about pre-grasp configurations with respect to the intended tasks. Further, a mapping is learned from part-related visual features to good grasping points. The probabilistic logic module makes use of object-task affordances and object/task ontologies to encode rules that generalize over similar object parts and object/task categories. The use of probabilistic logic for task-dependent grasping contrasts with current approaches that usually learn direct mappings from visual perceptions to task-dependent grasping points. We show the benefits of the full probabilistic logic 
pipeline experimentally and on a real robot.

\end{abstract}

\begin{IEEEkeywords}
task-dependent robot grasping, probabilistic logic pipeline, semantic grasping, local shape grasping
\end{IEEEkeywords}

\IEEEpeerreviewmaketitle
%%%%%%%%%%%%%%%%%%%%%%%%%%%%%%%%%%%%%%%%%%%%%%%%%%%%%%%%%%%%%%%%%%%%%%%%%%%%%%%%
\section{Introduction}

%by Laura
While robot vision capabilities are essential for perceiving and interpreting the world, robot grasping skills are essential for acting in arbitrary and dynamic environments and executing object manipulation tasks. Objects can be grasped in different ways. While performing a grasp, we must, at least, satisfy the grasping stability criteria, performing a good grasp also depends on the specific manipulation scenario: the object, its properties and functionalities, as well as task constraints and grasp constraints (e.g., gripper configuration). How to take into account such information for grasping is exactly the question we tackle in the present paper.

Specifically, instead of just learning a function that directly maps visual perceptions to task-dependent grasps, we introduce, as key contribution, an intermediary \emph{probabilistic logic} module to \emph{semantically reason} about the most likely object part to be grasped, given the object properties and task constraints. Then, a mapping is learned from part-related local visual features to good grasping points. The introduced symbolic part-based representation has several advantages:

\begin{itemize}
 \item grasp transfer to novel objects that share similar parts and thus, generalization over similar (multiple) object parts;
 \item high-level task-dependent reasoning over parts reduces possible grasps and hence, improves performance;
 \item the use of symbolic parts as manifold information for reliable object category estimation.
\end{itemize}

The present paper investigates robot grasping by leveraging symbolic world knowledge, in the form of object/task ontologies and object-task affordances, visual features, object categorical and task-based information, by integrating them into one \emph{probabilistic logic pipeline}. World knowledge and relations are encoded in compact logical grasping models that generalize over similar object and task categories, thus the logic offers a natural way to encode high-level knowledge. %This is a critical aspect of intelligence. 
However, often, descriptions of the perceived world are also uncertain. For example, not all cups look like the `prototypical' cup. Thus, we need probabilistic models, which, additionally, allow one to reason about the uncertainty in the world. We assume we have observations about the task and visual scene available and we show that, using our probabilistic logic, we can ask queries about different grasping aspects.

This paper makes the following contributions: 
\begin{enumerate}
 \item the integration of object category and task information for semantic object part grasping, 
 \item a first probabilistic logic module for task-dependent reasoning about robot grasping,
 \item a general rule-based model encoding object-task affordances and objects/tasks ontologies that reflect world knowledge and allow generalization over object/task categories.
\end{enumerate}

By employing object-task affordances and objects/tasks ontologies, the proposed pipeline can generalize over similar object parts and object/task categories. This allows us to experiment with a wide range of object and task categories, which is a critical aspect of autonomous agents acting freely in new environments. Our approach can be extended beyond the set of categories used, by augmenting the probabilistic logic module with extra rules. The benefit of introducing the proposed probabilistic logic pipeline is shown experimentally.

\subsection{The proposed pipeline}

\begin{figure}
 \begin{center}
  \includegraphics[width=0.45\textwidth]{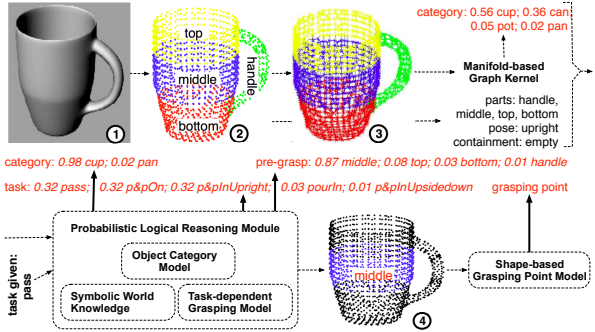}
 \end{center}
\caption{The task-dependent grasping pipeline on a \emph{cup} point cloud example. Top row (left to right): object $\textcircled{1}$, symbolic object parts $\textcircled{2}$ with labels \emph{top} (yellow), \emph{middle} (blue), \emph{bottom} (red), and \emph{handle} (green), $k$-nn graph $\textcircled{3}$ with part labels, $k=4$ (the edges are colored according to the colors of the adjacent nodes), manifolds model with its outcome and visual description of the object (pose, containment and parts). Bottom row: probabilistic logic module with its components and reasoning outcome, predicted pre-grasp \emph{middle} $\textcircled{4}$, shape-based grasping model and predicted grasping point.}
\label{fig:pipeline_example}
\end{figure}

The pipeline is exemplified in Fig.~\ref{fig:pipeline_example}. It takes as input the point cloud of an object (i.e., a cup) and, using vision-based techniques, we first obtain a description of the scene in terms of symbolic object parts, object pose and containment. We assess global object similarity via manifold-based graph kernels to complete the scene description with a prior on the object category. Next, using the visual description, we query the probabilistic logic module for the most likely object category, most likely task and best pre-grasp. For our cup example, the manifold shape model predicts the categories $cup$, $can$ and $pot$ with probabilities $0.56$, $0.36$ and $0.05$, respectively. The categorical logic module reasons about the symbolic parts and recalculates the probabilities as following: $0.98$ for category $cup$ and $0.02$ for category $pan$. The presence of exactly one \emph{handle} increases the probability of the object being more a \emph{cup} rather than a \emph{can} and 
identifies the object more as a \emph{pan} rather than a \emph{pot}. Similarly, using object-task affordance knowledge and object/task ontologies (e.g., any object affords the tasks \emph{pass} and \emph{pick-place on}), but also world knowledge (e.g., the task \emph{pour in} cannot be executed on a full object), the probabilistic logic module predicts the tasks \emph{pass}, \emph{pick-place on}  and \emph{pick-place inside upright} with equal probability. If the task given is \emph{pass}, the task-dependent grasping model predicts as most likely pre-grasp the \emph{middle} part of the object. The last step in the pipeline is the local shape-based grasping module which predicts the best point for grasping in the pre-grasp point cloud.

%%%%%%%%%%%%%%%%%%%%%%%%%%%%%%%%%%%%%%%%%%%%%%%%%%%%%%%%%%%%%%%%%%%%%%%%%%%%%%%%

\begin{figure*}[ht]
 \begin{center}
  \includegraphics[width=0.9\textwidth]{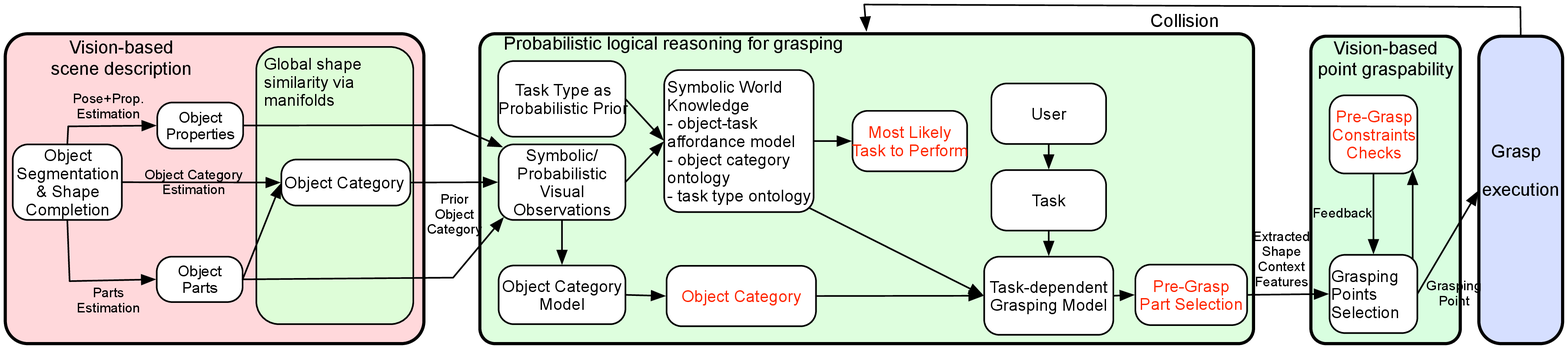}
 \end{center}
\caption{The task-dependent grasping pipeline.}
\label{fig:pipeline}
\end{figure*}

An overview of our learning and reasoning grasping pipeline is shown in Fig.~\ref{fig:pipeline}. It has four modules. The first module is a visual perception module (detailed in Section \ref{sec:module1}) which maps the object point cloud to scene descriptions in the form of symbolic and probabilistic visual observations about the world. After it segments the object point cloud and performs a full object shape reconstruction, the visual module estimates the object pose (\emph{upright}, \emph{sideways}) and parts (\emph{top}, \emph{middle}, \emph{bottom}, \emph{handle}, and \emph{usable area}). Further, it predicts a prior on the object category by employing object similarity based on manifold and semantic part information via a graph kernel. The second module is responsible for pre-grasp prediction. It consists of a probabilistic logic reasoning model for task-dependent grasping which, given the input observations, is able to perform inference about the grasping scenario. The module can answer, in turn, 
queries about the object category, most likely task and most likely object part to grasp, given the task. It uses probabilistic visual observations about the object, such as object pose, object category (i.e., a prior on the object category predicted by the aforementioned graph kernel-based probabilistic estimator), and object functionality (i.e., the object is empty or full) provided by the first module, and evidence about the task, either given by a user, or in the form of a probabilistic prior on the task type. A more detailed explanation of this module is provided in Section \ref{sec:module2}. Once we have identified the most likely object part to grasp, the pipeline calls the third module, which solves the problem of grasping point prediction using local shape features of the object part. The description of this module follows in Section \ref{sec:module3}. Finally, the last module takes the best predicted grasp and executes it on the robotic platform.

We proceed as follows. We start by reviewing the related work. Next, we explain in more detail each component of the proposed pipeline: the vision-based module, the probabilistic logic module and the grasping point prediction. Before concluding, we present our experiments in simulation as well as on a real robot.

%%%%%%%%%%%%%%%%%%%%%%%%%%%%%%%%%%%%%%%%%%%%%%%%%%%%%%%%%%%%%%%%%%%%%%%%%%%%%%%%

\section{Related Work}
%% by Laura

\subsection{Visual-dependent grasping}

The majority of grasping methods consider mainly visual features to learn a mapping from 2D/3D features to grasping 
parameters~\cite{Bohg2010362,DBLP:journals/ras/MontesanoL12,DBLP:journals/corr/abs13013592,Saxena08learninggrasp}. 
Nevertheless, these methods have a major shortcoming: it is difficult to link  a 3D gripper orientation to solely 
local image features. Only recently, methods that take global information into account have been 
proposed~\cite{DBLP:conf/icra/AleottiC11,Neumann12mlg}. The benefit is an increased geometric robustness, which is advantageous 
with respect to the pre-shape of the robotic hand and the general shape of the object, generating
more accurate grasps. However, global information relies on a complete shape of the object.

In recent years shape completion using a single view has been extensively studied, typically in robotics grasping applications. Usually multiple object partial views are acquired from different viewpoints, using 3D range cameras, and the gathered point clouds are then registered and aligned together in a common reference frame. The Iterative Closest Point algorithm~\cite{BeslICP1992} and efficient variants~\cite{RusinkiewiczICPvariants} are often used to compute the alignment transformations and to build a complete object shape model~\cite{ChenICP}. However, when only a single view is available and/or it is not possible to acquire several views due to time constraints or scenario/robot restrictions the shape completion problem becomes harder and some assumptions or pattern analysis must be made.
In this direction, a wide range of ideas have been proposed including fitting the visible object surface with primitive shapes such as cylinders, cones, parallelepipeds~\cite{Marton09ISRR,Kuehnle2008} or with more complex parametric representations like superquadrics~\cite{Biegelbauer2007}. 

Closely related to our shape completion approach, Thrun and Wegbreit~\cite{ThrunW05} proposed a method based on the symmetry assumption. This method considers 5 basic and 3 composite types of symmetries that are organized in an efficient entailment hierarchy. It uses a probabilistic model to evaluate and decide which are the completed shapes, generated by a set of hypothesized symmetries, that best fit the object partial view. More recently Kroemer et al. \cite{Kroemer2012} proposed an extrusion-based completion approach that is able to deal with shapes that symmetry-based methods cannot handle. The method starts by detecting potential planes of symmetry by combining the Thrun and Wegbreit method with Mitra et al.'s fast voting scheme~\cite{Mitra2006}. Given a symmetry plane, an ICP algorithm is used to decide the extrusion transformation to be applied to the object partial point cloud. Despite the fact that these methods were shown to be robust to noise and were able to deal 
with a wide range of object classes, they are inherently complex in terms of computational effort and thus, not suitable in real-time. Nevertheless, to simplify this problem, one can take advantage of common scenario structures and objects properties that are usually found in daily environments. They mostly involve man-made objects that are typically symmetric and standing on top of planar surfaces. For example, Bohg et al.~\cite{Bohg2011} took advantage of the table-top assumption and the fact that many objects have a plane of reflection symmetry. Starting from the work of Thrun and Wegbreit~\cite{ThrunW05} and similar in spirit to Bohg et al.~\cite{Bohg2011}, we propose a new computationally efficient shape completion approach which translates a set of environmental assumptions into a set of approximations, allowing us to reconstruct the object point cloud in real-time, given a partial view of the object.

\subsection{Task-dependent grasping}
Since  grasping is highly correlated with the task to be performed on the object, a lot of recent work has focused on incorporating task constraints in robot grasping. This is mostly done by learning a direct mapping function between good grasps and geometrical and action constraints, action features and object attributes. A part of this work focuses on Bayesian network learning to integrate symbolic task goals and low-level continuous features such as object attributes, action properties and constraint features~\cite{madry2012ICRA_SPME,5649406}. The goal is to learn features of importance for grasping knowledge transfer. This work is extended to consider object categorical information as an additional feature to predict suitable task-dependent grasping constraints~\cite{DBLP:conf/icra/MadrySK12}. Further, Detry et al.~\cite{detry2013ICRA,detry2012a,detry2012b} identify grasp-predicting prototypical parts by which objects are usually grasped in similar ways. The discrete part-based representation allows 
robust grasping. Differently, in addition to the semantic parts, we also consider a task-dependent setting that uses probabilistic logic and world-knowledge to reason about best pre-grasps. Several approaches make use of object affordances for grasping. While in~\cite{DBLP:conf/humanoids/SweeneyG07} the authors employ estimated visual-based latent affordances, the work in~\cite{BarckHolst2009LearningGraspingAffordance} reasons about grasp selection by modeling affordance relations between objects, actions and effects using either a fully probabilistic setting or a rule-based ontology. In contrast, we employ a probabilistic logic-based approach which can generalize over similar object parts and several object categories and tasks.

Related to our probabilistic logic pipeline is the fully probabilistic one introduced in~\cite{bohg2012SYRACO}. It combines low-level features and Bayesian networks to obtain possible task-dependent grasps. Closely related is the semantical pipeline presented in~\cite{DBLP:conf/iros/DangA12}. It employs a semantic affordance map which relates gripper approach directions to particular tasks. However, we exploit additional object/task ontologies using probabilistic reasoning and leverage low-level learning and semantic reasoning. This allows us to experiment with a wide range of object categories.

%by Marion & Laura
%Statistical relational learning (SRL) \cite{Getoor07, deReadt08} is an emerging area of machine learning that 
%combines probabilistic reasoning and relational representations with many successful applications in 
%social network modeling, text mining, bioinformatics, and robotics, such as collaborative filtering, entity resolution, 
%or the prediction of protein metal bindings.
%As one of the most ambitions goals of robotics is the understanding of artificial and natural intelligence,  
%learning what is relevant for the behavior in natural environments is becoming a central topic in modern robotics research.

\subsection{SRL for robot grasping and other robotic tasks}
From a different point of view, probabilistic relational robotics is an emerging area within robotics. Building on statistical relational learning (SRL) and probabilistic robotics, it aims at endowing robots with a new level of robustness in real-world situations. We review some recent successful contributions of SRL to various robotic tasks. Probabilistic relational models have been used to integrate common-sense knowledge about the structure of the world to successfully accomplish search tasks in an efficient and reliable goal-directed manner \cite{HanheideGDHWPAGZ11}. Further, relational dependency networks have been exploited to learn statistical models of procedural task knowledge, using declarative structure capturing abstract knowledge about the task~\cite{HartGJ05}. The benefits of task abstraction were shown in~\cite{Winkler2012acs}, where the robot uses vague descriptions of objects, locations, and actions in combination with the belief state of a knowledge base for reasoning. The goal of this work 
is to robustly solve the planning task in a generalized pick and place scenario. Abstract knowledge representation and symbolic knowledge processing for formulating control decisions as inference tasks have proven powerful in autonomous robot control \cite{Tenorth09knowRob}. These decisions are sent as queries to a knowledge base. SRL techniques using Markov Logic Networks and Bayesian Logic Networks for object categorization from 3D data have been proposed in~\cite{DBLP:conf/iros/MartonRJKB09}.

In probabilistic planning, relational rules have been exploited for efficient and flexible 
decision-theoretic planning \cite{LangT10} and probabilistic inference has proven successful for integrating
motor control, planning, grasping and high-level reasoning \cite{ToussaintPLJ10}. 
In mobile robotics, relational navigation policies have been learned from example paths with 
relational Markov decision Processes \cite{CocoraKPBR06}.
In order to compute plans comprising sequences of actions and in turn be able to solve complex manipulation tasks, 
reasoning about actions on a symbolic level is incorporated into robot learning from demonstrations \cite{abdo12tampra}. 
Symbolic reasoning enables the robot to solve tasks that are more complex than the individual, demonstrated actions. 
In~\cite{DBLP:conf/ijcai/KulickTLL13} meaningful symbolic relational representations are used to solve sequential manipulation tasks in a goal-directed manner via active relational reinforcement learning. Relational Markov networks have been extended to build relational object maps for mobile robots in order to enable reasoning about hierarchies of objects and spatial relationships amongst them \cite{LimketkaiLF05}.Related work for generalizing over doors and handles using SRL has been proposed in \cite{395061}.

All of these approaches successfully intertwine relational reasoning and learning in robotics. 
However, none of these frameworks solves the generalization capability needed for task-dependent 
grasping following an affordance-based behavior. Relational affordance models for robots have been 
learned in a multi-object manipulation task context \cite{349410}. 
We propose a probabilistic logic pipeline to infer pre-grasp configurations using object-task affordances.
%% by rui 
Dealing with unknown objects is an important  research topic in the field of robotics. This is because in
applications related to object grasping and manipulation, robots aimed at working in daily environments have to interact with many never-seen-before objects and increasingly complex scenarios.
%%%%%%%%%%%%%%%%%%%%%%%%%%%%%%%%%%%%%%%%%%%%%%%%%%%%%%%%%%%%%%%%%

\section{Vision-based Scene Description}
\label{sec:module1}
%Dealing with unknown objects is a strong research topic in the field of robotics.
%In applications related with object grasping and manipulation, robots aimed at working in daily environments have to interact with many never-seen-before objects and increasingly complex scenarios.

%Several object representations have been proposed and used in the past \todo{Citation!} to plan grasping and manipulation actions: %based on human recordings:
%complete meshes for known objects, reconstructed meshes for unknown
%objects, and object part clusters modeled with superquadrics, also for unknown objects. Despite these representations allowing for good precision 
%in representing the shape of the objects, they suffer from high-dimensionality and varying
%description length, thus being hard to define a representation of object categories suitable for generalization. 
%Further, the noise present in the sensor data may negatively influence representations with large numbers of parameters. The most robust and simple features to represent objects, match their similarity with others, and provide a basis for the definition of categories, must be low-dimensional and rely on gross features. 
%\todo{Not sure if the paragraph above should be integrated in the related work section, subsection VI A.}

The role of the visual module (cf. first module box in Fig.~\ref{fig:pipeline}) is to obtain a semantic description of the perceived objects in terms of their pose, symbolic parts and probability distributions over possible object categories. The object segmentation step~\cite{MujaTableTopURL} is followed by part detection and object category estimation, which rely on a full object point cloud. When only a partial view of the object is available, we employ a symmetry-based methodology for object shape completion. Next, the extraction of semantical parts is based on the object's dimensions along the main geometrical axes and can be achieved by bounding-box analysis via PCA. The low dimensional and efficient representation obtained guides the division of each object into a set of semantical parts, namely, \emph{top}, \emph{middle}, \emph{bottom}, \emph{handle} and \emph{usable area}. This reduces the search space for robot grasp generation, prediction and planning.
%When the complete object surface is available, global shape characteristics can be extracted. We use the main geometrical axes to define a local reference frame
%and a 3D bounding-box to define its size.
%the definition of object categories, thus being suitable for generalization. 
%todo{here a brief overview of what followes should be given. In general, we should try to create a better flow for the reader, 
%by connecting the sections and subsections better and thus explaining why we need the respective steps!}
The next subsections explain our symmetry-based method for shape-completion and the division of the completed point cloud into a set of semantical parts.

\subsection{Object shape completion}
\label{sec:point_cloud_completion}

As any other type of reconstruction based on single views, computing the bounding-box of the object is an ill posed
problem due to lack of observability of the self-occluded part. Thus, as for grasping procedures it is necessary that the robot knows the complete shape of the object of interest, some assumptions about the occluded part must be made. 
%We consider that objects present at least 180 degree symmetries, so that we are able to reconstruct the unobserved part of the point cloud. This models perfectlysimple object shapes like boxes, spheres and cylinders, and is a reasonable assumption for many objects of daily usage when lying on a table. 
Inspired by the work of Thrun and Wegbreit~\cite{ThrunW05} and Bohg et al.~\cite{Bohg2011} and with computational efficiency in mind, we propose a new approach that translates a set of assumptions and rules of thumb observed in many daily environments into a set of heuristics and approximations. They allow us to reconstruct the unobserved part of an object point cloud in real-time, given a partial view.

We consider the following assumptions: 
% which are true for a large number of objects and scenarios that we can find on a daily basis in household environments:
%\begin{enumerate}
% \item The objects stand on top of a planar surface (tabletop assumption) and the camera is standing on a higher viewpoint.
% \item The objects have rotational symmetry.
% \item The objects main geometrical axes are either orthogonal or parallel to the supporting plane.
% \item The axis of symmetry corresponds to one of the main geometrical axes.
% \item The direction of the axis of symmetry provides the object pose (upright or sideways).
% %\item The axis of symmetry is orthogonal or parallel to the supporting plane. % 
%\end{enumerate}
a) the objects stand on top of a planar surface (table top); b) the camera is at a higher viewpoint; c) the objects have rotational symmetry; d) their main geometrical axes are either orthogonal or parallel to the supporting plane; e) the axis of symmetry corresponds to one of the main geometrical axes; and f) the direction of the axis of symmetry indicates the object's pose (i.e., upright or sideways).

\begin{figure}[top]
\centering
%\subfigure[]{\includegraphics[width=.6\columnwidth]{x-yView}}
\includegraphics[width=.3\columnwidth]{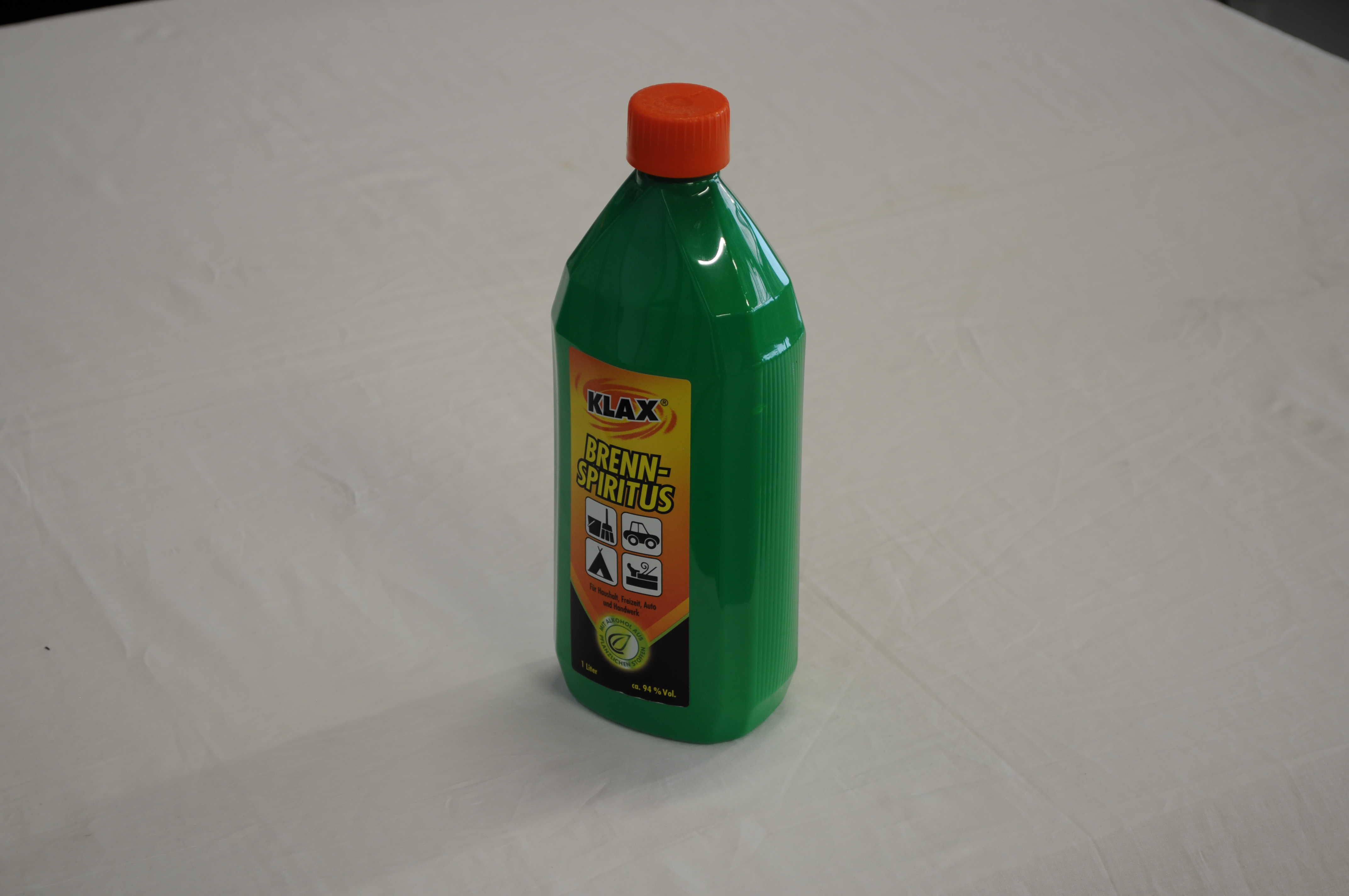}
\includegraphics[width=.3\columnwidth]{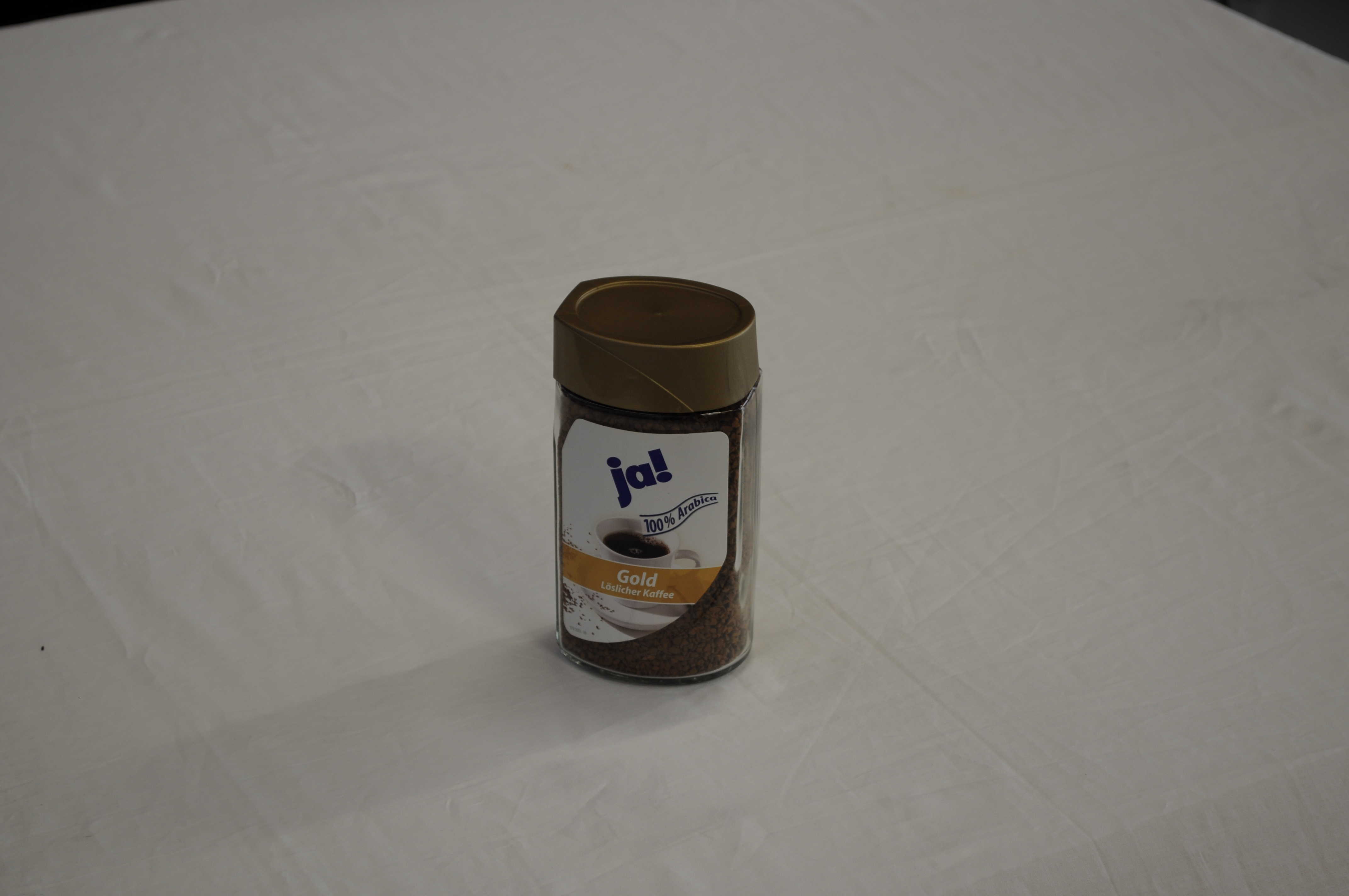}
\includegraphics[width=.3\columnwidth]{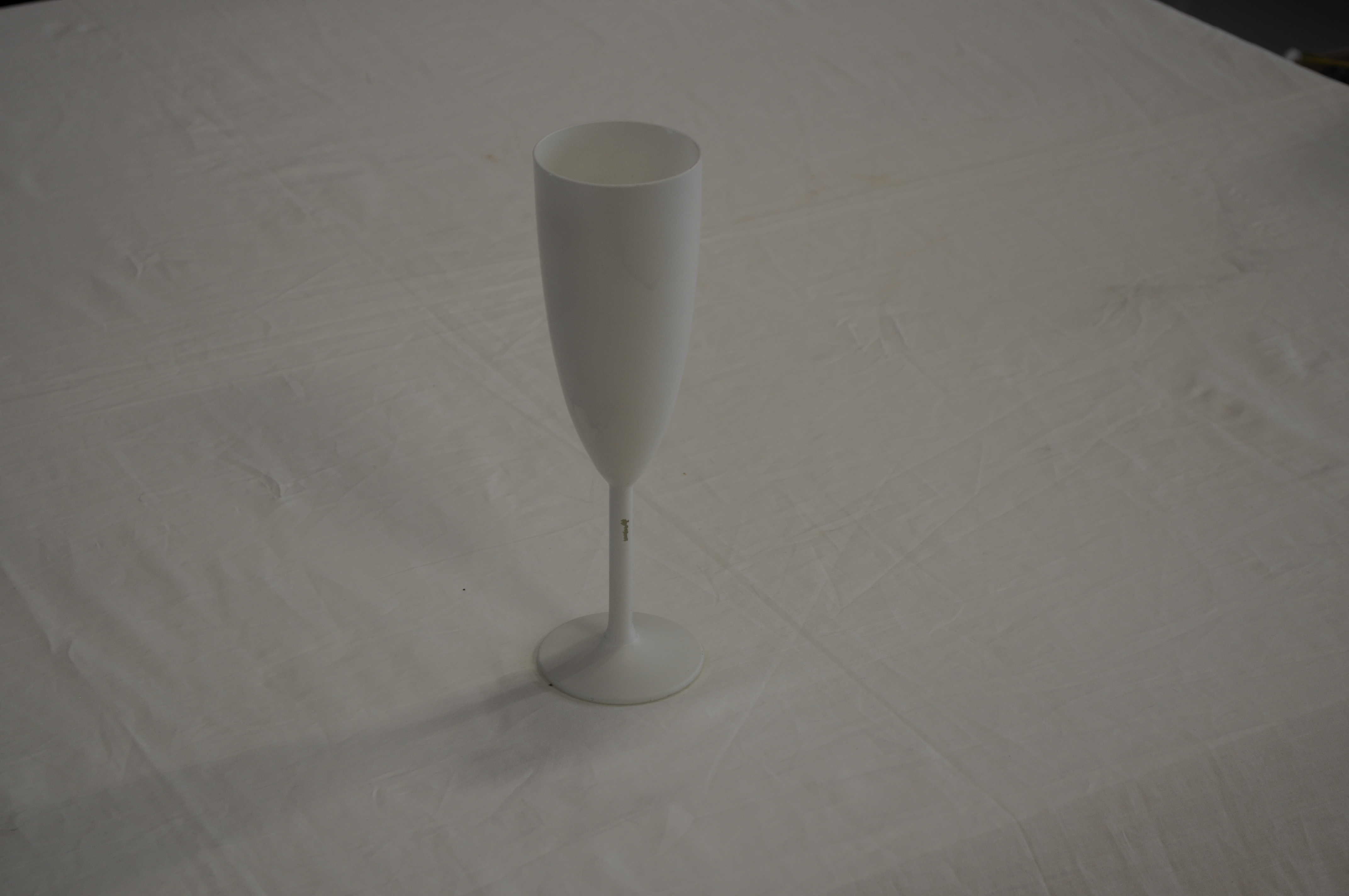}
\par\medskip
\vspace{-0.1cm}
\includegraphics[width=.3\columnwidth]{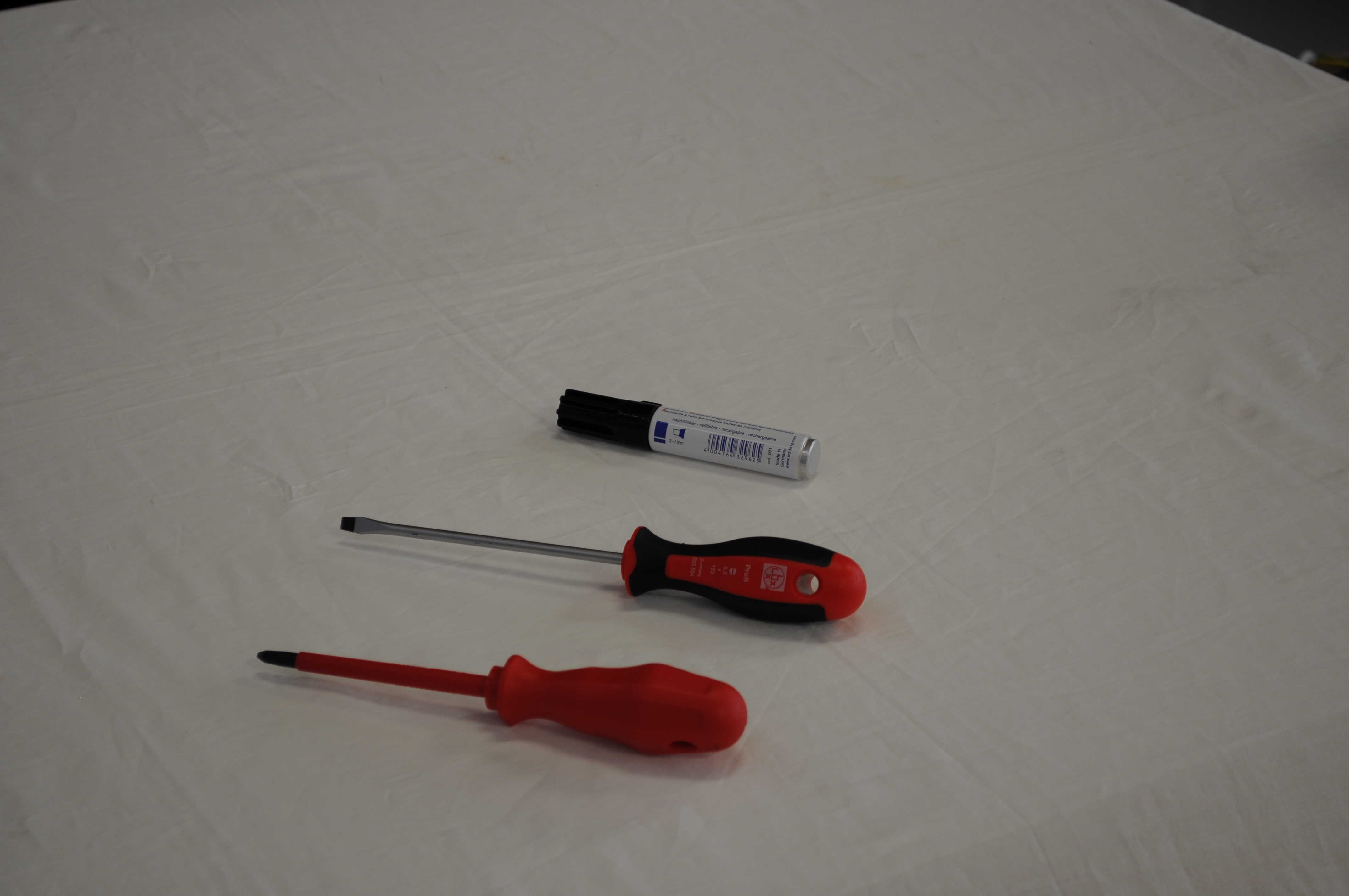}
\includegraphics[width=.3\columnwidth]{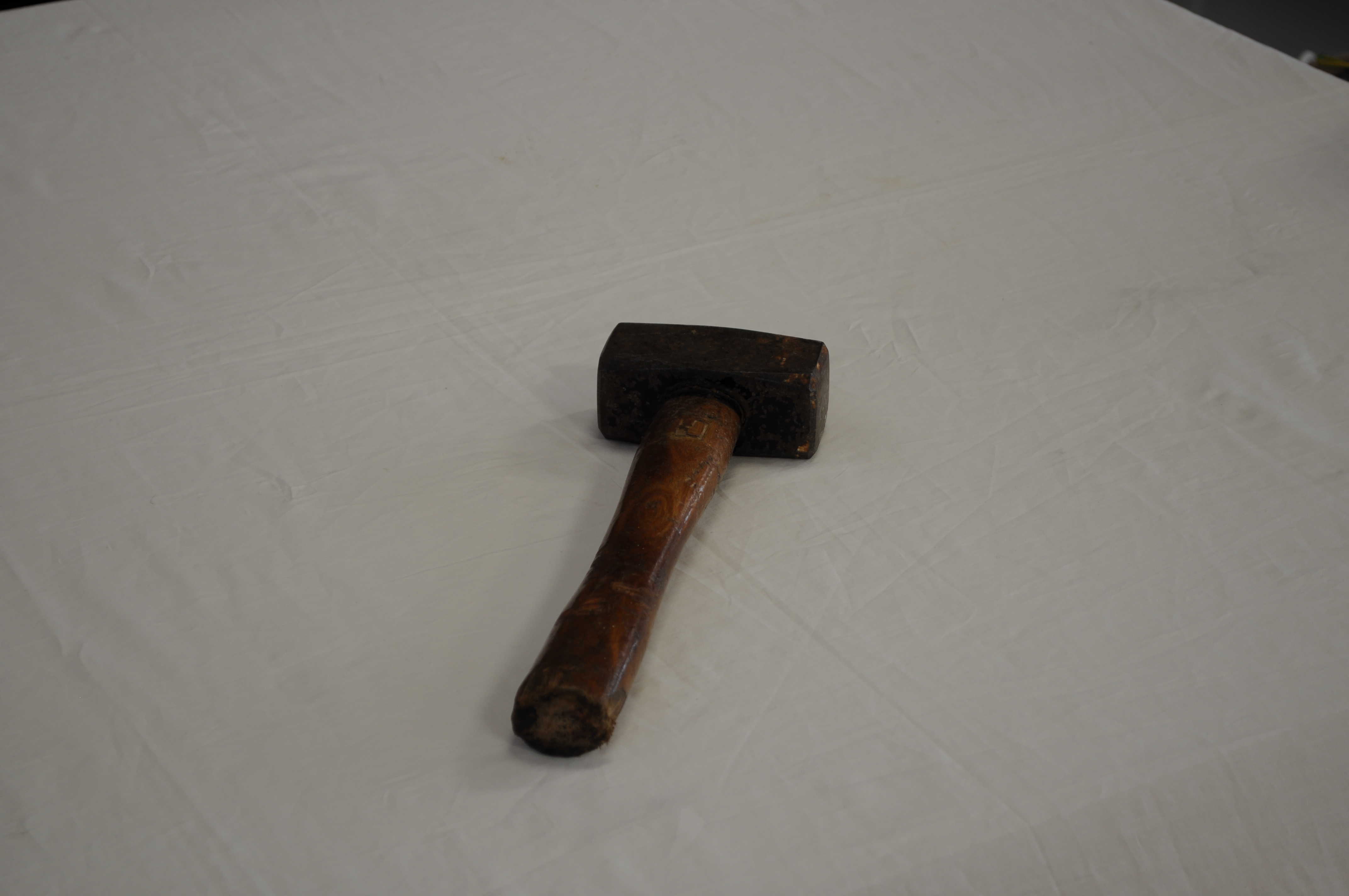}
\caption{Objects having rotational symmetry.}
\label{fig:objectSamples}
\end{figure}

These constraints model perfectly simple box-like and cylinder-like object shapes, such as kitchen-ware tools, and are
reasonable assumptions for many other approximately symmetric objects, such as tools (see Fig.~\ref{fig:objectSamples}). 
%of daily usage when standing or lying on a planar surface.
Analogous to~\cite{Bohg2011}, we consider only one type of symmetry, however we employ the \textit{line reflection symmetry}~\cite{ThrunW05} as it copes better with the object categories that we want to detect.
%\todo{verify the changes in this sentence again!}
%Our shape completion method consists on finding the object symmetry axis  follows: %and takes as input a segmented point cloud in the plane local reference frame, which is provided by a preceding planar segmentation algorithm.

\begin{figure}[top]
\centering
  \subfloat[]
 {
    \def\svgwidth{0.4\columnwidth}
    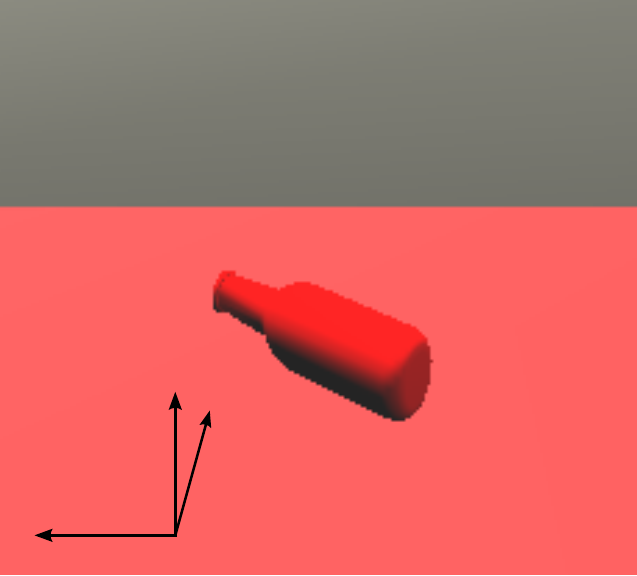
 }
 \subfloat[]
 {
    \def\svgwidth{0.5\columnwidth}
    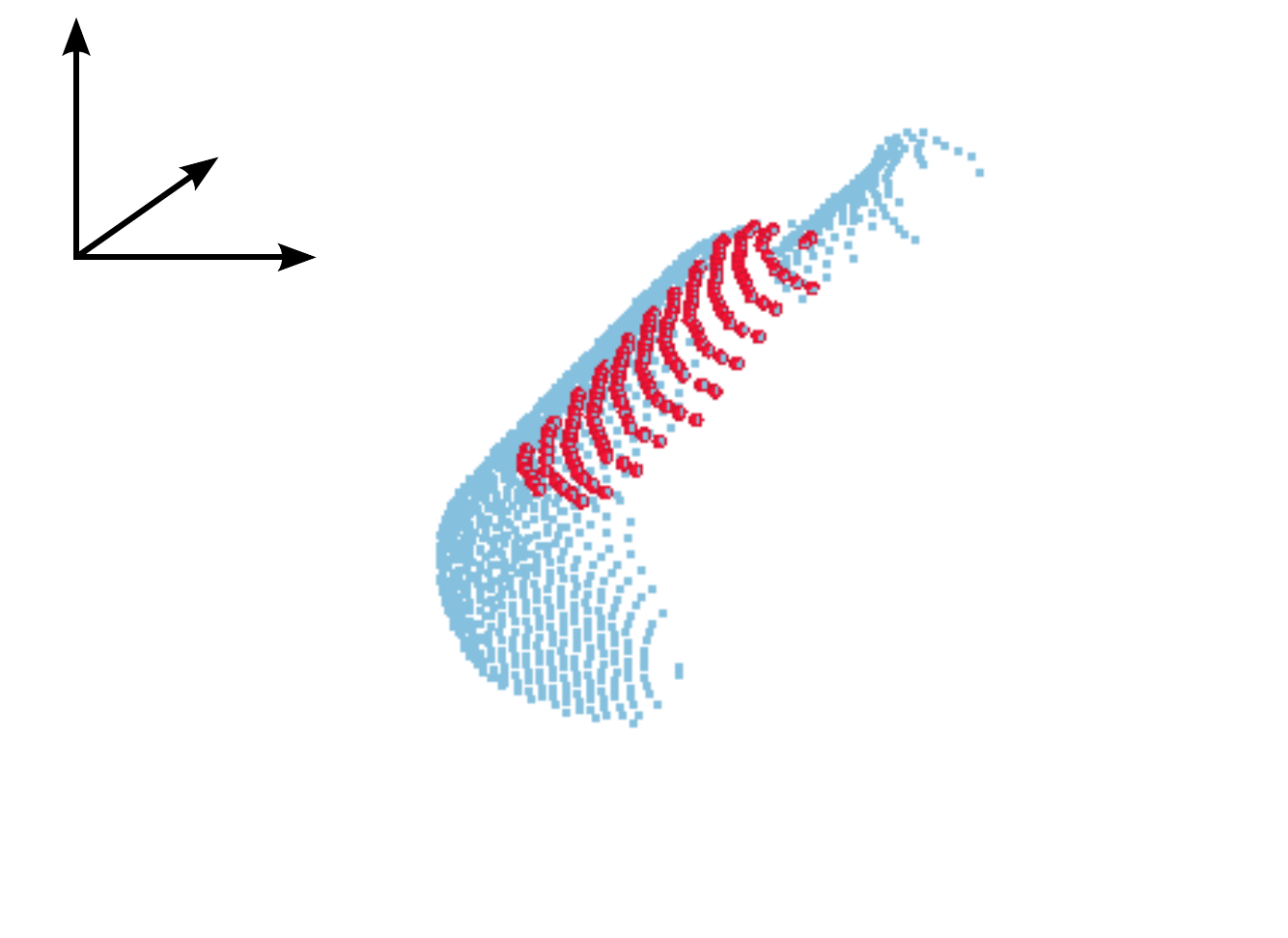
 }\\
\subfloat[]
 {
    \def\svgwidth{0.4\columnwidth}
    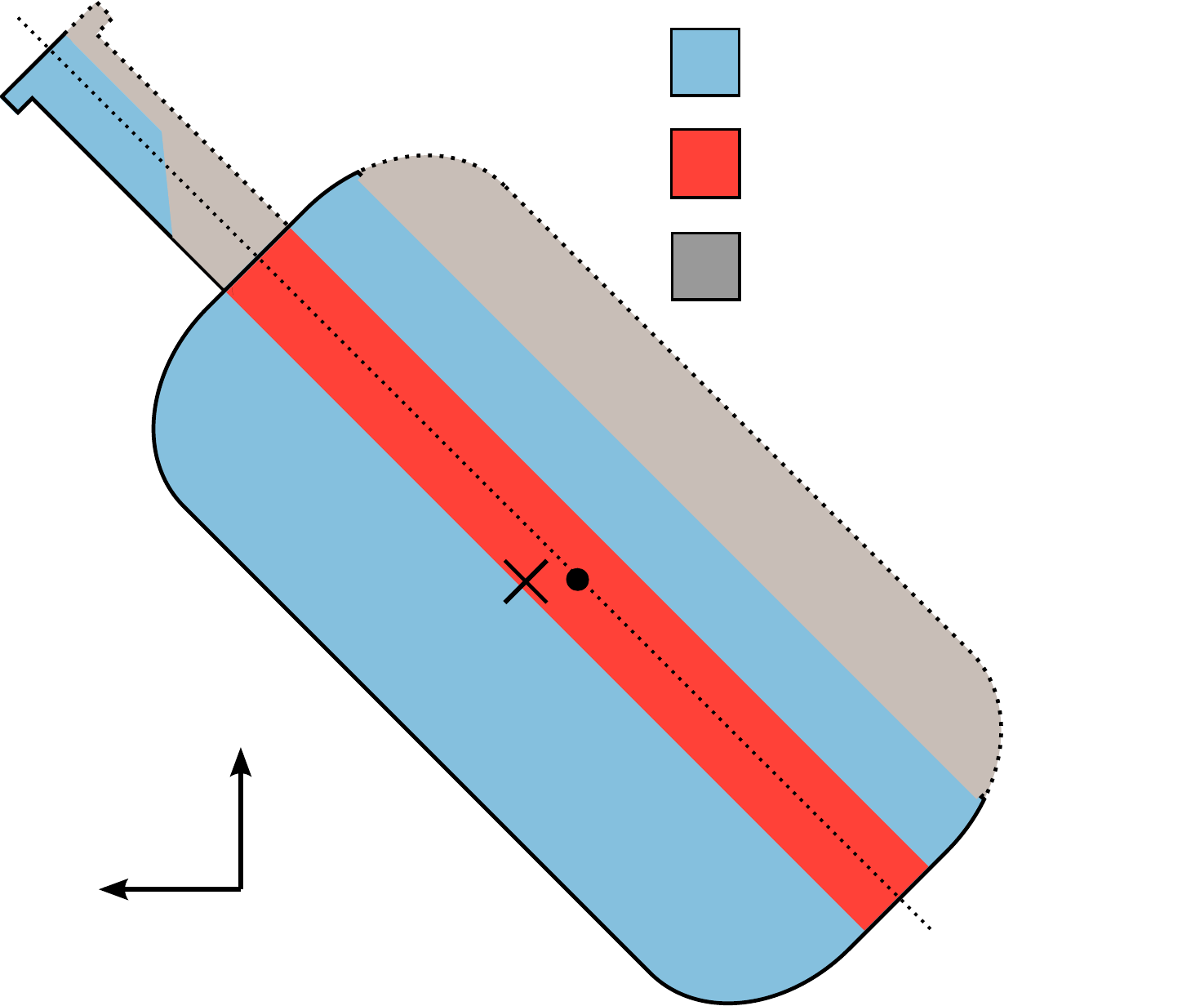
 }
 \subfloat[]
 {
    \def\svgwidth{0.5\columnwidth}
    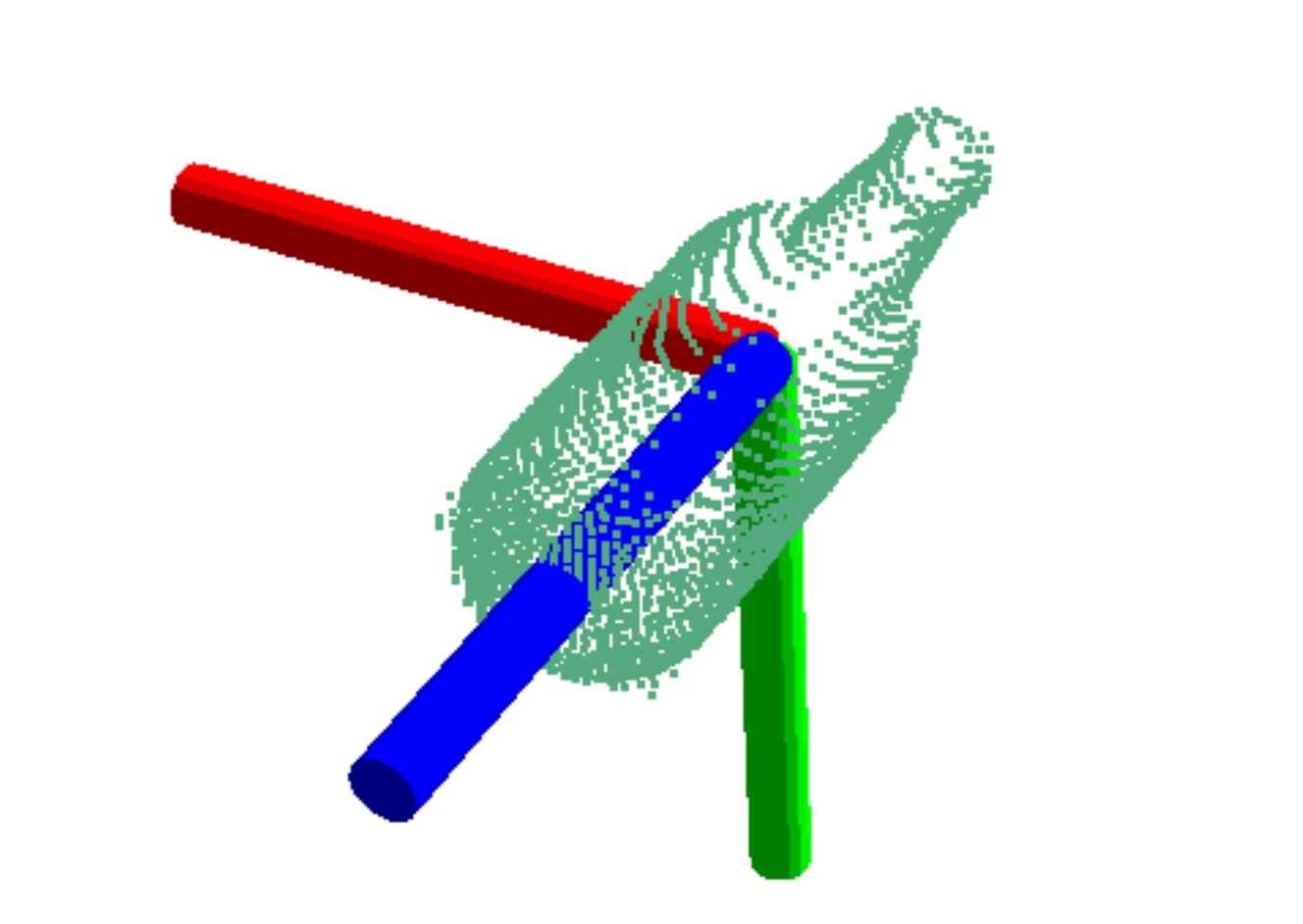
 }

\caption{2D centroid estimation in the presence of self-occlusion. (a) Bottle camera-view. (b) Visible region (blue) and top visible region surface points (red). (c) Bottle planar projection: $\times$ marks the centroid of $W$ (blue), whereas $\bullet$ indicates the centroid of $W_\text{top}$ (red). (d) After shape completion, an object coordinate frame is defined as having its origin at the bounding box centroid and $z$-axis aligned with the symmetry axis.}
\label{fig:perception1}
\end{figure}

Let $P=\{\mathbf{p}\}\subset\mathbb{R}^3$ be the set of visible object surface points. Our shape completion algorithm finds the object symmetry axis $s$ reflecting all visible surface points across it. This corresponds to rotating $P$ around $s$ by $180^\circ$.
%Let $Q=\{\mathbf{q}\}\subset\mathbb{R}^3$  be the set of surface points that approximates the hidden object region obtained by rotating each $\mathbf{p}$ around $s$.
%As previously mentioned we assume that the object axis of symmetry is either perpendicular or parallel to the supporting plane which means that the object is either standing or lying. %we first need to decide whether the object is standing or lying. 
We determine $s$ by analyzing the box  that encloses the set $P$, considering the principal directions of the box and the dimensions along those directions. The symmetry axis (i.e., principal direction) is orthogonal to the cross product of the bounding box directions whose dimensions are the closest, and passes through the bounding-box centroid.
%The symmetry axis direction is thus given by the direction that corresponds to the largest bounding-box edge of $P$
%or to the smallest if the two largest directions are similar (\todo figure needed). % do not differ more than a threshold $\delta$ 
To cope with the supporting plane assumption, we compute the horizontal (i.e., table plane, $xy$) and vertical (i.e. table normal, $z$) bounding-box directions and their dimensions separately. 
The vertical direction of the bounding-box is given by the normal vector to the table plane and its length is given by the furthest point from the supporting plane $d_z=\text{max}_{z}(P)$. Since the horizontal directions are arbitrarily oriented in the supporting plane, we apply the projection of $P$ onto the table plane and compute the directions and their dimensions in that space. The 2D components of the centroid location on the table plane cannot be correctly estimated from a partial view in most of the cases (as illustrated in Fig.~\ref{fig:perception1}). Let $W=\{\mathbf{p}_{z=0}\}\subset\mathbb{R}^2$ be the set containing the projected points. We assume that the top part of the object is visible and holds the symmetry assumptions so that the object's $xy$-centroid, $c_{xy}$, is obtained by considering only the top region points $W_\text{top}=\{\mathbf{w}_\text{top}\}\subset W$, satisfying the condition:%, i.e, %$p_z>\sigma  
\begin{equation}
 \mathbf{w}_\text{top}^i=\left\{ \begin{array}{ll}
       \mathbf{p}^i_{z=0} & \mbox{if $p_z>\sigma d_z$}\\
       \emptyset & \mbox{otherwise}\end{array} \right.
\end{equation}
where $\sigma\in[0,1]$ is a parameter tuned according to the camera view-point and the object shape curvature (i.e., $\sigma$ is higher for cylinder-like shapes and lower for parallelepiped-like ones).
 %we compute the transformation that aligns $W$ with the $xy$-plane reference frame with the view to obtain its horizontal bounding-box dimensions $\mathbf{d}_W=(d_x,d_y)$.
%Thereafter we compute the transformation $\mathbf{T}_{w\rightarrow t}$ that aligns $W$ with the table reference frame $t$, as follows:
%\begin{equation}
%\label{alignment_transformation}
%\mathbf{T}_{s\rightarrow t}=\mathbf{R}_{W}(P-\mathbf{c_{\text{top}}})
%\end{equation}
%with $\mathbf{R}_W$ denoting the rotation matrix that aligns the eigenvectors $\mathbf{v}_1$ and $\mathbf{v}_2$ of $W$ with the table reference frame $x$ and $y$ axis and $\mathbf{c}_{\text{top}}$ the $P$ top region projection centroid.
%After aligning $W$ with $t$ we determine the projection bounding-box edges sizes associated with the eigenvectors $\mathbf{v}_1$ and $\mathbf{v}_2$.
The eigenvectors provided by PCA on the set $W$ define the horizontal directions whereas their lengths are given by projecting the points in $W$ onto its eigenvectors and finding the maximum in each direction.

%2D bounding-box center used to compute the axis of rotational symmetry and the principal axis of the projected cloud are used to estimate the rotation to be applied.
 %The algorithm is based on finding the object plane of symmetry and rotating the point cloud around it 180 degrees. Fig. 1 shows some examples with the results of this process.
\subsection{Part-based object representation}
\label{sec:partdetection}
\begin{figure}[t]
\centering
\subfloat[Pan]{\includegraphics[width=.3\columnwidth]{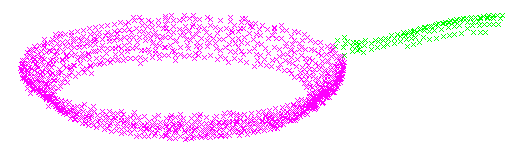}}
\subfloat[Knife]{\includegraphics[width=.3\columnwidth]{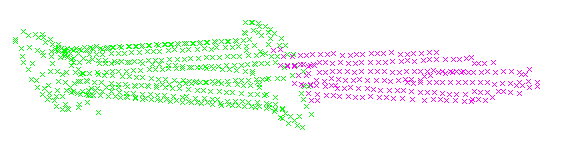}}
\subfloat[Hammer]{\includegraphics[width=.3\columnwidth]{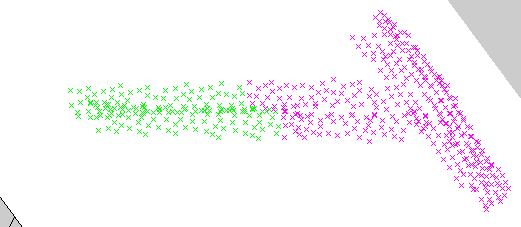}}\\
\subfloat[Glass]{\includegraphics[width=.15\columnwidth]{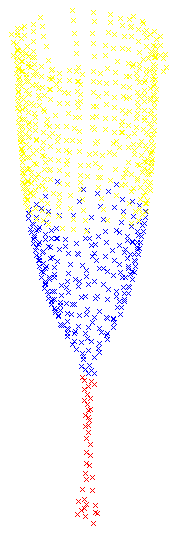}}
\subfloat[Bowl]{\includegraphics[width=.25\columnwidth]{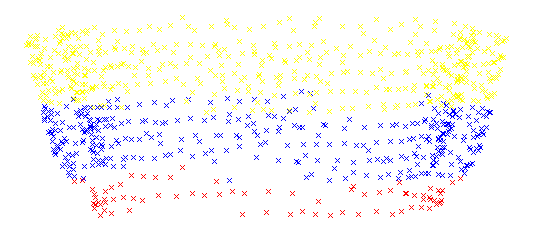}}
\subfloat[Mug]{\includegraphics[width=.3\columnwidth]{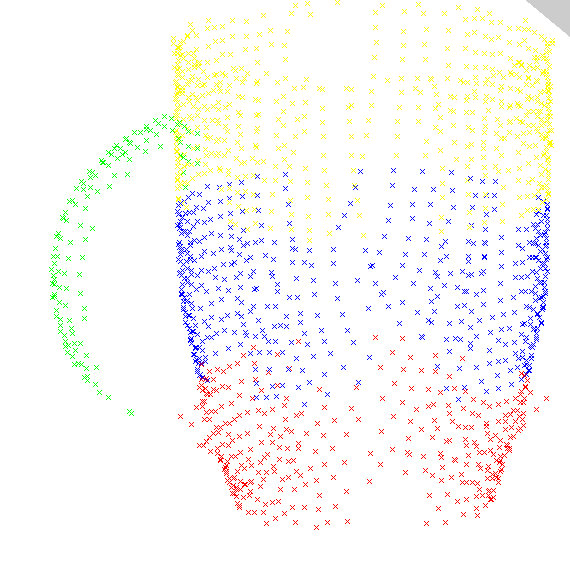}}
\caption{Semantic parts for several objects after applying the completion algorithm. The colors correspond to parts as follows: yellow - top, blue - middle, red - bottom, green - handle, and magenta - usable area.}
\label{fig:partSamples}
\end{figure}

%Taking this into account we adopted bounding-boxes to represent and delimit object semantical parts.
%According to the object main geometrical axis inter-relationships we compute the
We consider two main types of objects: tools and other objects. A tool has as parts a \emph{handle} and a \emph{usable area}, while the rest of the objects have \emph{top}, \emph{middle}, \emph{bottom} parts and may have \emph{handles}. When the axis of symmetry is parallel to the supporting plane and the lengths of the remaining directions are smaller than a predefined threshold, we consider that the object has a \emph{handle} and a \emph{usable area}. In order to cope with objects such as mugs and pans we detect a handle if a circle is fitted in the projected points $W$ with a large confidence. The points lying outside of the circle are labeled as \emph{handle}. The rest of the points are divided along the axis of symmetry into \emph{top}, \emph{middle} and \emph{bottom}. Fig.~\ref{fig:partSamples} illustrates examples of detected semantic parts for several objects using our completion algorithm. 

\begin{figure}[t]
\begin{center}
\includegraphics[width=.408\columnwidth, height=4.5cm]{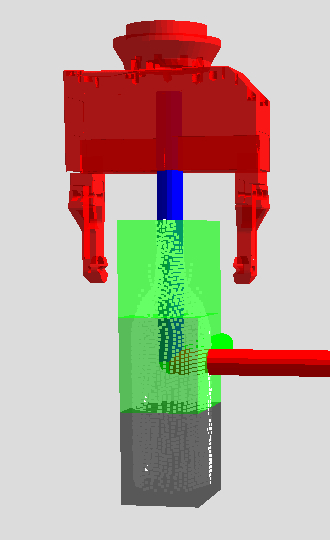}
\includegraphics[width=.3\columnwidth,height=4.5cm]{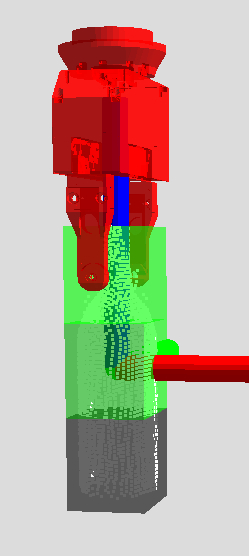}
\end{center}
\caption{Examples of the pre-grasp gripper poses for a face of the top part of a bottle.}
\label{fig:preGraspSamples}
\end{figure}

The bounding boxes of the object parts define the pre-grasp hypotheses, providing two pre-grasp poses for each face of a box, as illustrated in Fig.~\ref{fig:preGraspSamples}. The final number of pre-grasp hypotheses is pruned in a first stage by the task-dependent logical module and in a second stage by a collision checker and the motion trajectory planner.

\subsection{Object category via manifolds}
\label{sec:manifolds}
Given the completed object point cloud and its semantic parts, we estimate the object category by retrieving the objects with the most similar global properties. This manifold- and part-based category classification approach was introduced in~\cite{Neumann13mlg}.
%This manifold- and part-based category classification is one of the main contributions of this paper. 
While beneficial for direct grasping point prediction~\cite{Neumann12mlg}, global object similarity also ensures a strong enough appearance-based predictor for object category. This prediction in the form of a distribution on object categories is used as a prior for the probabilistic logic module. To incorporate global object similarity for object category prediction, we leverage propagation kernels, a recently introduced graph kernel designed for classification and retrieval of partially labeled graphs~\cite{NeumannPGK12}.

We obtain the distribution on object categories for a particular query object by retrieving the objects in an object database being most similar in terms of global shape and semantic part information. We represent the objects by labeled graphs, where the labels are the semantic part labels derived by the visual module and the graph structure given by a $k$-nearest neighbor ($k$-nn) graph. For each object point cloud we derive a weighted $k$-nn graph by connecting the $k$ nearest points w.r.t.\ Euclidean distance in 3D. 
%Accordingly, the edge weights for edge $(i,j)$ are set to the inverse of the Euclidean distance $w_{i,j} = \Vert\bold{x}_i - \bold{x}_j\Vert ^ {-1}$. 
We use a four-neighborhood (i.e., $k=4$) and assign an edge weight reflecting the tangent plane orientations of its incident nodes to encode changes in the object surface. The weight of edge $(i,j)$ between two nodes is given by $w_{i,j} = |\mathbf{n}_i \cdot \mathbf{n}_j|$, where $\mathbf{n}_i$ is the normal of point~$i$. The nodes have five semantic classes encoding object part information \emph{top}, \emph{middle}, \emph{bottom}, \emph{handle} and \emph{usable area}. To be able to capture manifold information as graph features in presence of full label information we use a diffusion scheme of the labels corresponding to the \emph{diffusion graph kernel}, in the following simply referred to as \emph{propagation kernel}, proposed in~\cite{NeumannPGK12}. We stress that the graphs of the 3D point clouds as illustrated\footnote{The illustration does not depict the edge weights being proportional to the change in curvature of the adjacent points.} in Fig.~\ref{fig:pipeline_example}$\textcircled{3}$ capture 
both manifold information (geodesic distance) via their structure and semantic information (part labels) via their node labels. 

The similarity measure among objects is a kernel function over counts of similar node label distributions per diffusion iteration.
The $T$-iteration propagation kernel between two graphs $G'$ and $G''$ is, then, defined as:
\begin{equation}
% G_t^{(i)} = (V^{(i)},E^{(i)},L_t^{(i)}), \; t = 0,...,T 
  K_T(G', G'') = \sum_{t = 0}^T ker(G'_t, G''_t), 
\label{equ:prop_kernel}
\end{equation}
where $T$ represents the maximum number of label propagation interactions considered and $ker$ is a linear base kernel, defined as:
\begin{equation}
  ker(G'_{t}, G''_{t}) = \langle\phi(G'_{t}),\phi(G''_{t})\rangle.
\label{equ:base_kernel}
\end{equation}
In our experiments we vary $T \in \{0,...,15\}$ and use the maximum number of iterations giving the best results. 
 
The main ingredients of propagation kernels are the distribution-based graph features $\phi(G_{t})$. 
They are essentially computed from node label distributions of running label diffusion 
%as introduced in \cite{ZhuGL03} 
on the respective graphs. Hence, the node label distributions of $G_{t}$ are updated according to 
$L_{t}  \leftarrow T \, L_{t-1}$, where the transition matrix $T$ is the 
row-normalized adjacency matrix $T = D^{-1} A$ and $D$ being the diagonal degree matrix with $D_{ii} = \sum_j A_{ij}$. 
Based on the $L_{t} (0 \leq t  \leq T)$, we compute for each graph the counts of similar distributions among the respective graphs' nodes.
% Let $\ell^{(i)}_{t,j}$ be the $j$-th row of $L^{(i)}_t$ and 
% $\mathcal{L} = \bigcup_i^N \bigcup_j^{n_i} \lbrace \ell^{(i)}_{t,j} \rbrace$ be the set of all uniquely occurring 
% label distributions on the nodes of all $N$ graphs in the database and $n_i$ is the number of nodes in graph $G_i$.
% Then 
% \begin{align}
% \phi(G^{(i)}_t) = \sum_{j = 1}^{n_i} f(\ell^{(i)}_{t,j}),
% \label{equ:feature}
% \end{align} 
% where $f$ is a mapping from the space of distributions $\mathbb{R}^k$ into the space of standard basis vectors 
% $E_{k'} = \left\lbrace e_1,...,e_{k'} \right\rbrace$ with $k' = |\mathcal{L}| \leq n$, 
% where $n$ is the number of nodes for all graphs $n=\sum_{i=1}^N n_i$. 
As the node label distributions are $m-$dimensional \emph{continuous} vectors, where $m$ is number of semantic labels, in this case $m=5$, 
propagation kernels use locality sensitive hashing (LSH)~\cite{Datar04} as a quantization function to ensure the acquisition of meaningful features. We employ a quantization function for distributions preserving the total variation distance (for more details, see~\cite{NeumannPGK12}). The bin width parameter of LSH is fixed to $w=10^{-4}$ in all experiments.

Propagation kernels leverage the power of evolving continuous node label distributions as graph features and hence, 
capture both manifold information and semantic information. 
Given a new object $G^*$ that the robot aims to grasp, we first select the top $n$ most similar graphs $\{G^{(1)}, \cdots, G^{(n)}\}$, 
where the similarity is given by the respective row of the correlation matrix $\hat K$, where $\hat K_{ij} = \frac{K_{ij}}{\sqrt{K_{ii} K_{jj}}}$. 
Note, that by using  $\hat K$ instead of $K$ we achieve a normalization w.r.t.\ the number of points in the point clouds and hence, w.r.t.\ 
the scale of the objects. We set $n=10$ in all our experiments. Second, we build a weighted average over the categories of the objects corresponding 
to $\{G^{(1)}, \cdots, G^{(n)}\}$ where the weight function is defined as $f(x) = \exp{(-x)}$ with $x$ being the rank after sorting the kernel row $\hat K_{*,:}$.
This average is finally used as a prior distribution on the object category for the object with the graph representation $G^*$. The prior distribution over object categories for the cup in Fig.~\ref{fig:pipeline_example} is:

{\scriptsize
\begin{Verbatim}[commandchars=\\\{\},codes={\catcode`$=3\catcode`_=8}]
0.56 cup; 0.36 can; 0.05 pot; 0.02 pan
\end{Verbatim}
}
\noindent and will be used further by the probabilistic logic module to reason about the different prediction tasks.

\section{The Probabilistic Logic Module}
\label{sec:module2}
After the visual module, we introduce our reasoning module (cf. second module in Fig.~\ref{fig:pipeline}). Its role is to answers three types of queries, related to object category, most affordable task and best semantic pre-grasp. For example, for object category prediction we query the object instance for being a hammer by calculating the probability $P(cat(O,hammer)|obs,M)$, where $O$ is the object to classify, $M$ is the model and $obs$ is the conjunction of observations made about the world (e.g., $obs$ = \{parts, pose, task\}). Similarly, we query for the most likely pre-grasp by calculating probabilities $P(grasp(Pt)|obs,M)$, where $Pt$ is an object part. When the task is not observed the set of observations becomes $obs$ = \{parts, pose\} and then the task can be also inferred from the model by calculating probabilities $P(affords(O,T)|obs,M)$, where $T$ is a task.

Bayesian Networks (BNs) are often used to model such complex dependencies involving uncertainty~\cite{DBLP:conf/icra/MadrySK12}. Differently, our probabilistic model is defined using Causal Probabilistic (CP) Logic~\cite{DBLP:journals/tplp/VennekensDB09}. There are several advantages of a CP-Logic. First, it can intuitively integrate world knowledge as logic rules. For example, we can exploit object ontologies to reason about object (super-)categories. Similarly, we can use object/task ontologies and task-object affordance models to reason about possible, impossible and desirable pre-grasps. Second, CP-Logic is designed to explicitly model causal events (or relationships between random variables). For example, if the object has a usable area and a handle it is likely to be a `tool' and it can be one (any) of the tool sub-categories (e.g., `hammer'). This rule is a general, but local piece of information which does not consider other possible causes for the object sub-category. This is rather difficult to 
encode with a BN, as querying for $P(cat(O,hammer)|obs,M)$ involves knowing all the possible causes for $cat(O,hammer)$ and how they interact with the observations $obs$. Similarly, if the object is a `tool' and the task is `pass', then it should be rather grasped by the usable area instead of by  the handle. This involves again local causation. In fact, robotic grasping is characterized by a number of causal uncertain events which sometimes involve different consequences. Third, a CP-theory is more efficient as it requires fewer parameters~\cite{DBLP:journals/fuin/MeertSB08} and allows parameter sharing by generalizing over similar situations.

Our grasping CP-theory has 4 parts: 1) a set of probabilistic observations about the world consisting of visual object properties and, optionally, a probability distribution on the task type; 2) a set of logic rules, defined as background knowledge which incorporate common sense knowledge about the world, that is, object/task ontologies and object-task affordances; 3) a set of probabilistic logic rules as the object category model; and 4) a task-dependent grasping model in the form of probabilistic logic rules. Note that for all prediction tasks we make the mutually exclusiveness assumption. For object category prediction this implies that an object cannot have several categories at the same time. Similarly, for task selection, this translates into the fact that only one task can be executed at any point in time. We use a ProbLog implementation \cite{310668} of the CP-Logic theory and we show experimentally that by putting together probabilistic and logical reasoning we improve the grasping performance. We 
expose the numerated parts in the following subsections.

\subsection{Observations about the world}

In this paper, we observe one object at a time.\footnote{The work is easily extendable to consider several objects simultaneously.} \emph{Visual and task observations} of the scenario in Fig.~\ref{fig:pipeline_example} are shown below. They can be ground probabilistic facts, such as $0.8::part(top,o)$ stating that the object $o$ has a top part with probability $0.8$, deterministic facts, such as $empty(o)$ stating that $o$ is empty, or CP-rules. A CP-rule capturing the prior distribution over the object category for $object(o)$ is:

{\scriptsize
\begin{Verbatim}[commandchars=\\\{\},codes={\catcode`$=3\catcode`_=8}]
0.56::cup(o); 0.36::can(o); 0.05::pot(o); 0.02::pan(o)<- \\
   object(o).
\end{Verbatim}
}
\noindent stating that an object $o$ belongs to a category with a certain probability, that is, it is either a cup with probability $0.56$ or a can with probability $0.36$ or a pot with probability $0.05$ or a pan with probability $0.02$. In any CP-rule, the sum of the possible outcomes can be at most $1$.\footnote{If the sum is less than 1, there is a non-zero probability that nothing happens.} The probability should be interpreted as: if the body is true, then it causes the consequence to become true with a causal probability. Similarly, we can have a prior on the task type as a CP-event. In our experiments, if the task is not given, we assume a uniform distribution on the task type. Similarly, if the prior on the object category is not observed, we consider a uniform prior instead.

{\scriptsize
\begin{Verbatim}[commandchars=\\\{\},codes={\catcode`$=3\catcode`_=8}]
%visual observations for object o
object(o).
0.8::part(top,o). % the prob. that o has a top is 0.8
1.0::part(handle,o).
1.0::part(middle,o).
1.0::part(bottom,o).
0.5::pose(o,upright).
empty(o).
0.56::cup(o); 0.36::can(o); 0.05::pot(o); 0.02::pan(o)<- \\
   object(o).

%all tasks {$t_{1},\dots,t_{7}$} observed
pourOut($t_{1}$).
pass($t_{2}$).  ...
%prior on the task type
1/7::task($t_{1}$); 1/7::task($t_{2}$); ... ; 1/7::task($t_{7}$) <- true.
\end{Verbatim}
}
\noindent CP-rules with probability $1.0$ are encoded deterministically.

\subsection{World knowledge: ontologies and affordances}
\label{sec:worldknowledge}

\begin{figure}[t]
\begin{center}
	\includegraphics[width=8cm,height=4cm]{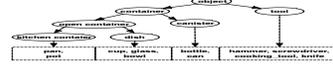}
	\caption{Object Ontology}
	\label{fig:ontoObj}
\end{center}
\end{figure}

The object ontology is illustrated in Fig.~\ref{fig:ontoObj} and structures $11$ object categories that we consider in our scenario: $C=\{$\emph{pan, pot, cup, glass, bowl, bottle, can, hammer, knife, screwdriver, cooking\_tool}$\}$. The super-categories, defined based on the object functionality, are: \emph{kitchenContainer, dish, openContainer, canister, container, tool, object}. By making use of the ontology structure, the grasping model makes abstraction of the fine-grained object categories.

\begin{figure}[t]
\begin{center}
	\includegraphics[width=8cm,height=3.5cm]{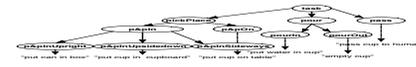}
	\caption{Task Ontology}
	\label{fig:ontoTask}
\end{center}
\end{figure}

The task ontology in Fig.~\ref{fig:ontoTask} structures $7$ tasks: $T=\{$\emph{pass, pourOut, pourIn, p\&pInUpright, p\&pInUpsidedown, p\&pInSideways, p\&pOn}$\}$. For example, the task \emph{pourOut} refers to the action of removing the contained liquid, while \emph{p\&pInUpsidedown} refers to picking and placing the object inside a shelf in the upside-down pose. Depending on the object properties, its parts and the task to be performed, the object should be grasped in different ways.

\begin{figure}[t]
\begin{center}
	\includegraphics[width=8.5cm,height=4cm]{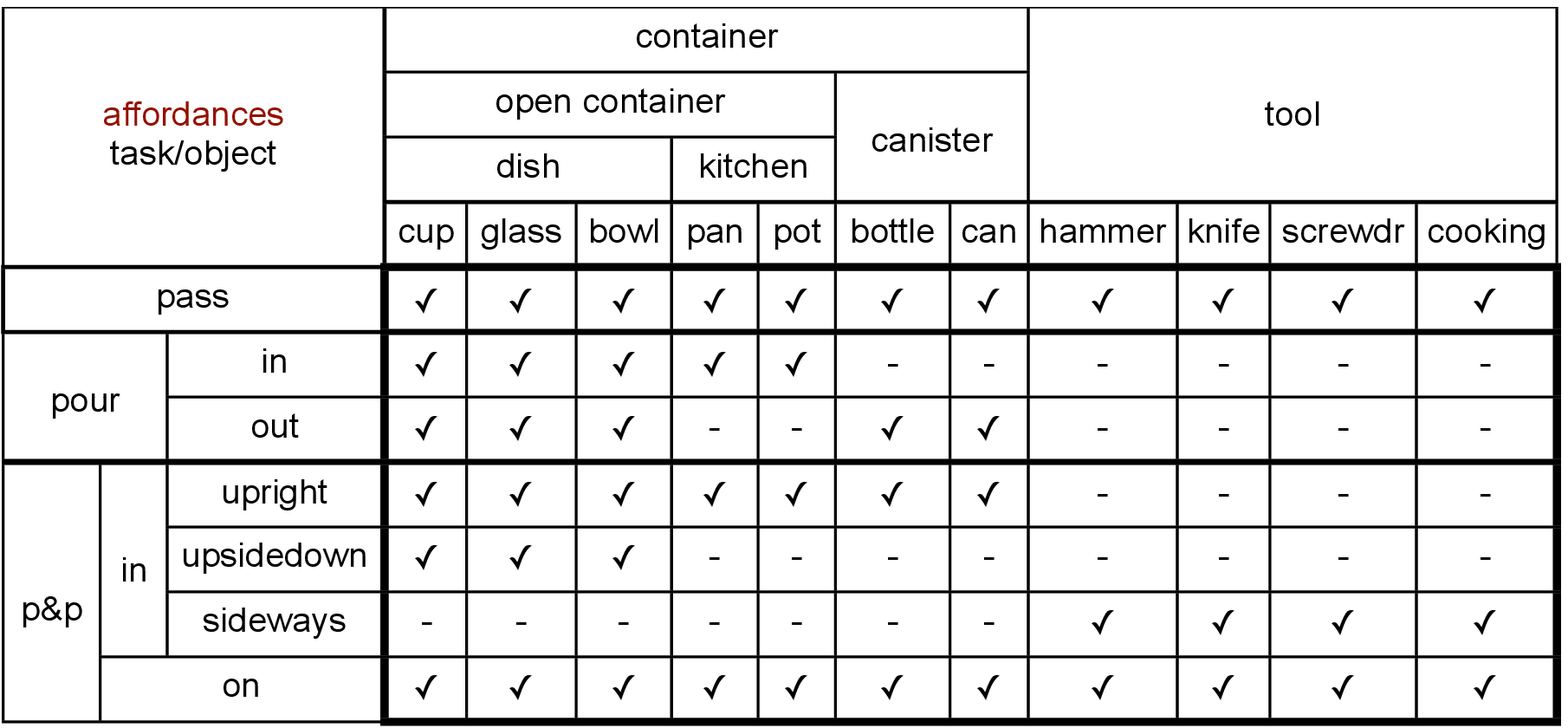}
	\caption[]{Object-Task Affordances}
\label{fig:aff}
\end{center}
\end{figure}

The object-task affordances for the considered scenario are illustrated in Fig.~\ref{fig:aff}. They allow us to relate the two concepts and thus define the grasping model in a relational way. Both the affordances table and the object ontology were defined by human experience and inspired by \emph{AfNet: The Affordance Network}\footnote{Available at: \url{www.theaffordances.net}.}. They can be extended to include new object/task categories.

This knowledge is translated into  deterministic logical rules\footnote{For convenience, in the case of task prediction, we use the affordance rules with a high equal probability due to the mutually exclusiveness constraint.} in the following way:

{\scriptsize
\begin{alltt}
%Examples of object ontology mappings
canister(X):- can(X).
dish(X):- cup(X).
tool(X):- hammer(X).
container(X):- canister(X).
object(X):- tool(X).
object(X):- container(X).
% Examples of task ontology mappings
pour(T):- pourIn(T).
pour(T):- pourOut(T).
task(T):- pour(T).
% Task affordances
possible(X,T):- object(X), pass(T).
possible(X,T):- container(X), pour(T).
possible(X,T):- object(X), p&pOn(T).
possible(X,T):- container(X), p&pIn(T).
possible(X,T):- tool(X), p&pInSideways(T).
% Some impossible task affordances
impossible(X,T):- canister(X), pourIn(T).
impossible(X,T):- kitchenContainer(X), pourOut(T).
% Common sense exceptions
impossible(X,T):- pan(X), full(X), pass(T).
impossible(X,T):- container(X), full(X), p&pInUpsidedown(T).
...
affords(X,T):- possible(X,T), not(impossible(X,T)).	
\end{alltt}
}
\noindent where, for example, $dish(X):- cup(X)$ is a deterministic intensional rule stating that ``any cup is a dish"; $dish(X)$ is the head of the rule, while $cup(X)$ is the body. We initially assume a deterministic affordances model. However, the model can be learned to obtain better estimations~\cite{349410} or can be re-estimated in our reasoning module by inferring the most likely task.  This shows the flexibility of our approach.

\subsection{The CP-theory for semantic grasping}
\label{sec:semanticgrasping}
Similarly, we use deterministic rules and CP-events indicating object category consequences based on the object parts and properties. For example the deterministic rule:

{\scriptsize
\begin{alltt}
tool(X):- part(ua,X),part(ha,X),pose(X,sideways).
\end{alltt}
}
\noindent reads as: if the object has a usable area and a handle and it poses sideways, then it is a tool. When observed parts are, for example, $bottom$, $middle$ and $top$, no handles are detected and the pose is sideways or upright, then the object can be a glass, a bowl or a canister. If the observed pose is upsidedown then the object can be a glass, a bowl or a can. If exactly one handle is observed, then the object may be a cup or a pan. In these cases we can define CP-events showing the possible outcomes:

{\scriptsize
\begin{alltt}
0.25::glass(X); 0.25::bowl(X); 0.5::canister(X) <- 
    part(top,X), part(middle,X), part(bottom,X), 
    no_handle(X), pose(X,upright).
0.25::glass(X); 0.25::bowl(X); 0.5::canister(X) <- 
    part(top,X), part(middle,X), part(bottom,X), 
    no_handle(X), pose(X,sideways).
0.33::glass(X) ; 0.33::bowl(X) ; 0.33::can(X) <- 
    part(top,X), part(middle,X), part(bottom,X), 
    no_handle(X), pose(X,upsidedown).
0.75::cup(X); 0.25::pan(X) <- part(middle,X),
    part(top,X), part(bottom,X), part(handle,X),
    pose(X,upright).
\end{alltt}
}

These rules encode generality also by using object super-category atoms in the head. Thus, in order to estimate the object category,  we replace the original object ontology defined in Section~\ref{sec:worldknowledge} with an ontology that models category distributions with respect to the super-categories across the ontology. This is part of the categorization model and is done using CP-events. The causal probabilities are estimated based on the number of specific categories in the leafs. For example, we have the distribution over \emph{hammer}, \emph{knife}, \emph{screwdriver} and \emph{kitchen\_tool} caused by the super-category \emph{tool} or the distribution over \emph{can} and \emph{bottle} caused by the object being a \emph{canister}:

{\scriptsize
\begin{alltt}
0.25::hammer(X); 0.25::knife(X); 0.25::screwdriver(X);
0.25::kitchen_tool(X) <- tool(X).
0.5::can(X); 0.5::bottle(X) <- canister(X).
\end{alltt}
}
\noindent Thus, in our experiments, the object categorization CP-theory $M$ contains category rules, world knowledge and visual observations. The CP-theory parameters should not be interpreted as the conditional probability of the head atom given the body, e.g., $P(can(o)|canister(o)) = 0.5$ is incorrect. It is part of the semantics of CP-Logic that each rule independently makes a head atom true when triggered. Thus, the conditional probability that $o$ is a can, given that $o$ is a canister, may be different than $0.5$, in case there is a second possible cause, e.g., a prior knowledge that $o$ is a can with probability $0.36$, which contributes to $P(can(o))$. 

\emph{Querying for the most likely object category} is equivalent to calculating $argmax_{_{C}} P(cat(o,C)|M)$. It is possible to ask the query without grounding the specific object category. This will result in a probability distribution over object categories, which is better than the observed prior. For the example in Fig.~\ref{fig:pipeline_example} the new distribution is $P(cat(o,cup))=0.98, P(cat(o,pan))=0.02$, while $P(cat(o,can))$ and $P(cat(o,pot))$ become $0$.

There are different levels of generalization with respect to the rules of the theory. We experimented also with more general rules to investigate the suitability of our model. A more general theory was also able to improve the object category prior (see section~\ref{sec:exp}), showing similar behavior and results to the more specific one. Examples of more general rules are shown below, where we replace, for example, the more specific head \emph{0.25::glass(X); 0.25::bowl(X); 0.5::canister(X)} with the super-category \emph{1.0::container(X)}, while keeping the same rule body. For the example in Fig.~\ref{fig:pipeline_example} the new distribution with the more general theory becomes \emph{P(cat(o,cup))=0.93, P(cat(o,pot))=0.05, P(cat(o,pan))=0.02}. 

{\scriptsize
\begin{alltt}
1.0::container(X) <- part(middle,X), part(top,X), 
    part(bottom,X), no_handle(X), pose(X,upright).
0.6::dish(X); 0.4::canister(X) <- part(top,X), part(middle,X),
    part(bottom,X),no_handle(X), pose(X,sideways).
1.0::container(X) <- part(middle,X), part(top,X), 
    part(bottom,X), no_handle(X), pose(X,upsidedown).
0.5::cup(X); 0.5::kitchen_container(X) <- part(middle,X), 
    part(top,X), part(bottom,X), one_handle(X).
\end{alltt}
}

To \emph{query for the most likely task} in a grasping scenario, we use world observations, the probability distribution over object categories and world knowledge. We can estimate the most likely task by calculating $argmax_{_{T}} P(affords(o,T)|M)$, where $M$ is the CP-theory. Again, one can ask queries without grounding $T$ to obtain a probability distribution over possible task types. For the example in Fig.~\ref{fig:pipeline_example} the distribution over possible tasks is \emph{P(affords(o,pass))=0.32, P(affords(o,p\&pOn))=0.32}, \emph{P(affords(o,p\&pInUpright))=0.32},\emph{P(affords(o,pourIn))=0.03}, \emph{P(affords(o,p\&pInUpsidedown))=0.01}.

We define the part-based grasping model as a set of CP-events. Each causal event generates as consequence the graspability of a certain object part conditioned on the part existence, task, object (super-)category and properties. The feasibility of the semantic grasp is encoded via the causal probability. Some examples from the grasping model for the \emph{dish} super-category are shown below:

{\scriptsize
\begin{alltt}
0.8::grasp(X,T,middle) <- affords(X,T), pass(T), dish(X), 
    pose(X,upright), full(X), part(middle,X).
0.1::grasp(X,T,top) <- affords(X,T), pass(T), dish(X),
    pose(X,upright), full(X), part(top,X).

0.1::grasp(X,T,bottom) <- affords(X,T), pass(T), dish(X),
    pose(X,upright), empty(X), part(bottom,X).
0.6::grasp(X,T,middle) <- affords(X,T), pass(T), dish(X),
    pose(X,upright), empty(X), part(middle,X).
0.2::grasp(X,T,top) <- affords(X,T), pass(T), dish(X), 
    pose(X,upright), empty(X), part(top,X).
0.1::grasp(X,T,handle) <- affords(X,T), pass(T), dish(X),
    pose(X,upright), full(X), part(handle,X).

0.2::grasp(X,T,bottom) <- affords(X,T), pass(T), dish(X), 
    pose(X,upsidedown), part(bottom,X).
0.7::grasp(X,T,middle) <- affords(X,T), pass(T), dish(X), 
    pose(X,upsidedown), part(middle,X).
0.1::grasp(X,T,top) <- affords(X,T), pass(T), dish(X), 
    pose(X,upsidedown), part(top,X).
...

0.7::grasp(X,T,middle) <- affords(X,T), p&pIn(T), dish(X), 
   pose(X,sideways), part(Id,middle,X).
0.3::grasp(X,T,bottom) <- affords(X,T), p&pIn(T), dish(X), 
   pose(X,sideways), part(bottom,X).

1.0::grasp(X,T,middle) <- affords(X,T), pourIn(T), dish(X), 
   empty(X), part(middle,X).
1.0::grasp(X,T,middle) <- affords(X,T), pourOut(T), dish(X), 
   not(empty(X)), part(middle,X).
...
\end{alltt}
}

We can enforce constraints to model impossible pre-grasps. For example, when the object is a $tool$ and the task is $pour$, we have an impossible affordance and thus, an impossible pre-grasp position. Examples of such constraint rules in ProbLog are:

{\scriptsize
\begin{alltt}
%constraint for impossible affordances
false:- grasp(X,T,R), task(T), object(X),
   impossible(X,T),part(R,X).
%constraint for collision
false:- grasp(X,T,R), task(T), object(X), 
   part(R,X), collision(R).
%other constraints
false:- grasp(X,T,R), pose(X,upsidedown),
   pan(X), task(T), part(R,X).
...
\end{alltt}
}

The first constraint states that it is impossible that the pre-grasp atom $grasp(X,T,R)$ is true when the body is true. This will guarantee that the probability of such grasps is equal to $0$. The second constraint rule shows that, additionally, we can connect the reasoning module to the execution planner by enforcing the probability of a pre-grasp to $0$ if there are environmental constraints for the gripper. The third constraint indicates that if the object is a pan in the upside down pose then no task should be executed, as grasping the object in this situation is very difficult.

If $M$ is the CP-theory for task-dependent grasping, we can \emph{query for the most likely semantic pre-grasp} of an object. This is equivalent to calculating $argmax_{_{Pt}} P(grasp(o,t_2,Pt)|M)$, where $Pt$ is a part in the set of observed object parts and $t_2$ is the given task. For the example in Fig.~\ref{fig:pipeline_example} the distribution over possible parts when the task considered is \emph{pass} becomes: \emph{P(grasp(o,pass,middle))=0.87}, \emph{P(grasp(o,pass,top))=0.08}, \emph{P(grasp(o,pass,bottom))=0.03}, \emph{P(grasp(o,pass,handle))=0.01}.

Similar to the object categorization theory, there are different levels of generalization with respect to the rules. To test the fittingness of the theory we experimented also with more general rules, by generalizing over the object pose and containment with respect to several tasks and thus, reducing the number of rules. For example, we replaced part of the theory presented above for task \emph{pass} and super-category \emph{dish} with:

{\scriptsize
\begin{alltt}
0.1::grasp(X,T,bottom) <- 
   affords(X,T), pass(T), dish(X), part(bottom,X).
0.6::grasp(X,T,middle) <- 
   affords(X,T), pass(T), dish(X), part(middle,X).
0.2::grasp(X,T,top) <- 
   affords(X,T), pass(T), dish(X), part(top,X).
0.1::grasp(X,T,handle) <- 
   affords(X,T), pass(T), dish(X), part(handle,X).
\end{alltt}
}

We have defined our models using human experience and ``educated guesses". They can be augmented by adding extra rules to include new object/task categories. The world knowledge was encoded as general as possible while still reflecting the ontologies and task-object affordances. The parameters of the rules composing the models can, in principle, be learned from data~\cite{DBLP:journals/fuin/MeertSB08} to best represent the application domain. Our current experimental results with the quite rigid affordance model can be improved by learning better probability estimates for object-task affordances from data. 

%%%%%%%%%%%%%%%%%%%%%%%%%%%%%%%%%%%%%%%%%%%%%%%%%%%%%%%%%%%%%%%%%%%%%%%%%%%%%%%%

\section{Local Shape Grasping Prediction}
\label{sec:module3}

%1. Reminder of why this module fits in the pipeline.
The probabilistic logic module selects the object part to be grasped and/or the task to be performed by taking into account high level information. Starting from here, the third module of the pipeline (cf. third module box in Fig.~\ref{fig:pipeline}) uses additional local shape features characterizing the object part to compute the grasping probability. It calculates $P(grasp | local \text{ }shape)$ by mapping the classification output of a Support Vector Machine (SVM) onto a probability. The SVM classifier discriminates between graspable and non-graspable shapes.

%3. Feature computation
\subsection{Depth difference features}
\begin{figure}[t]
 %\subfigure[]{\includegraphics[width=.6\columnwidth]{x-yView}}
\centering
\includegraphics[width=.35\columnwidth]{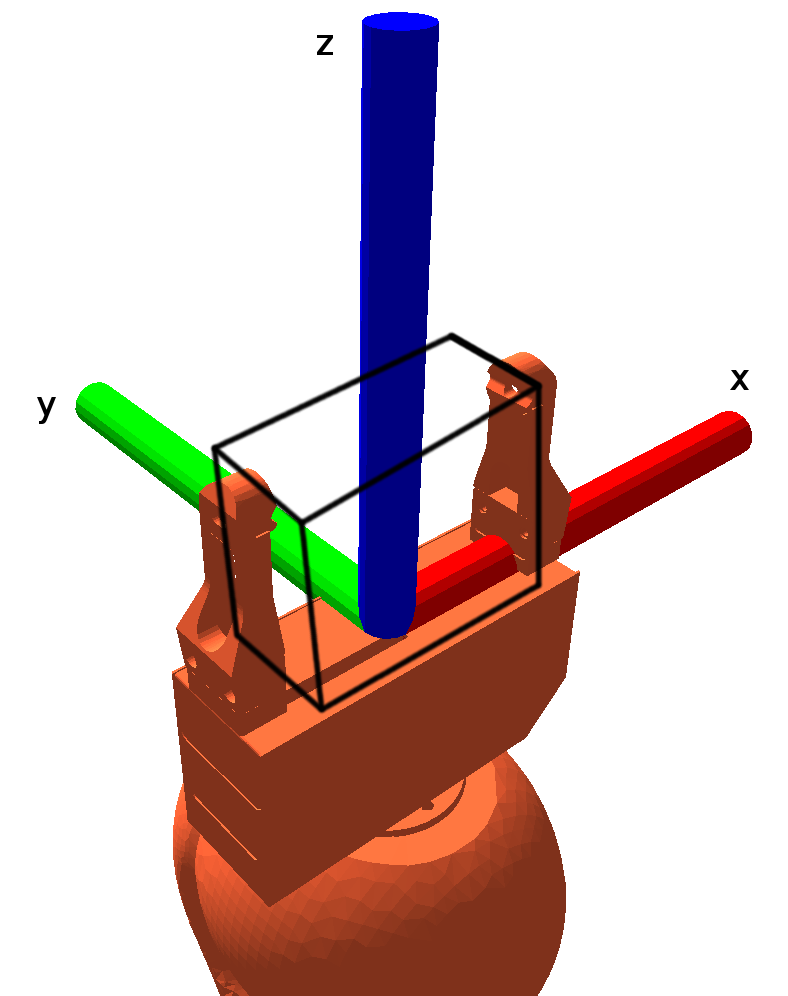}
\caption{Gripper and volume of interest, showing the reference frame origin for the orthogonal projection of the DI image from Eq. \eqref{eq:depthImage}.}
\label{fig:gripperEnclosingVolume}
\end{figure}

\begin{figure}[t]
\centering
 %\subfigure[]{\includegraphics[width=.6\columnwidth]{x-yView}}
\includegraphics[width=.5\columnwidth, height=3cm]{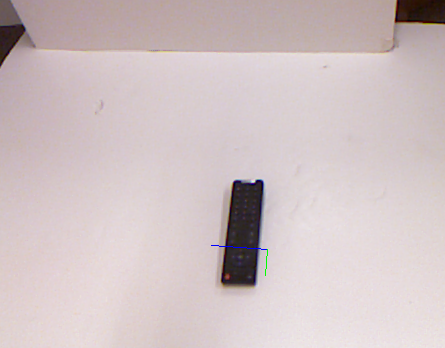}
\includegraphics[width=.45\columnwidth, height=3cm]{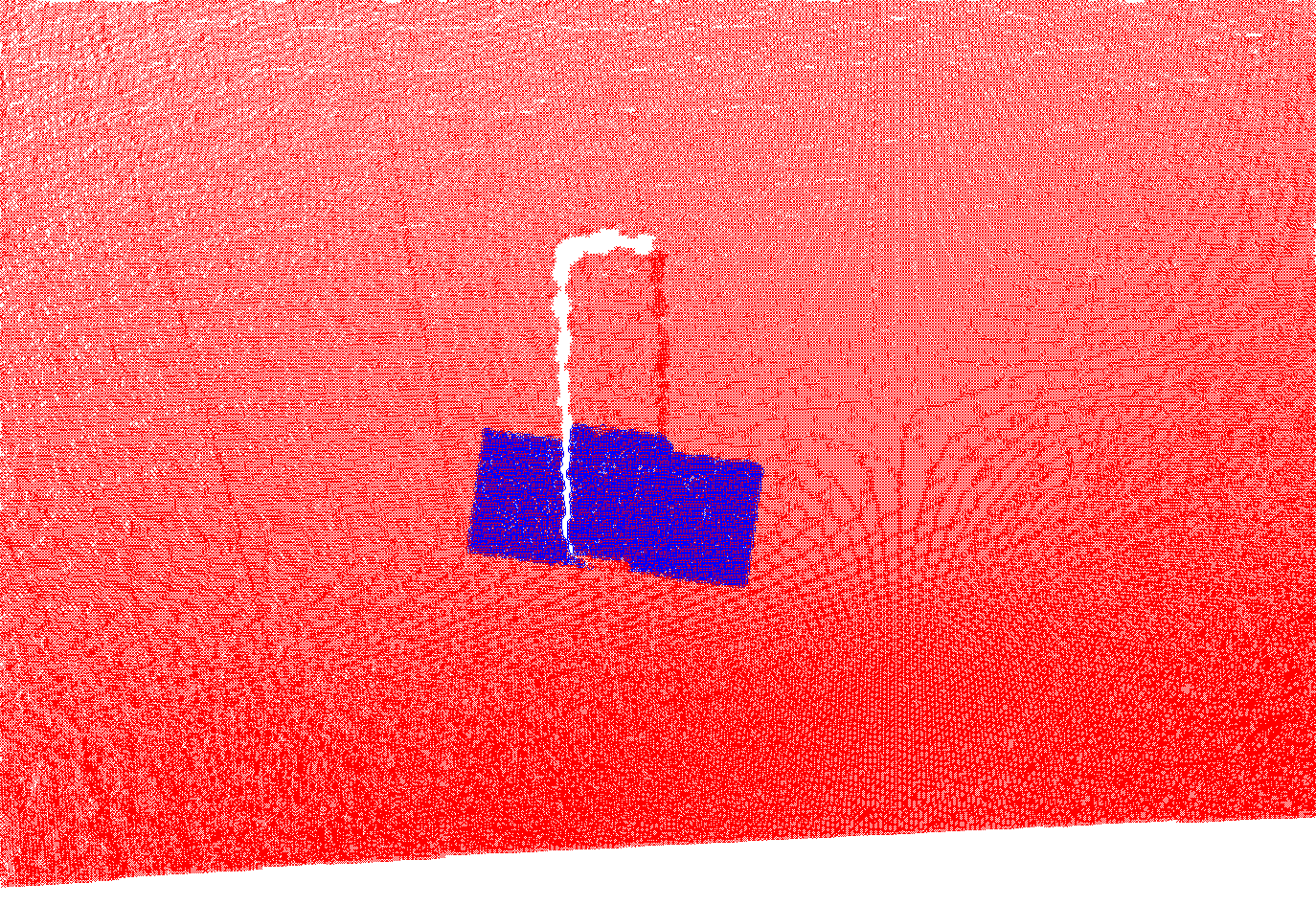}
\includegraphics[width=.7\columnwidth, height=3cm]{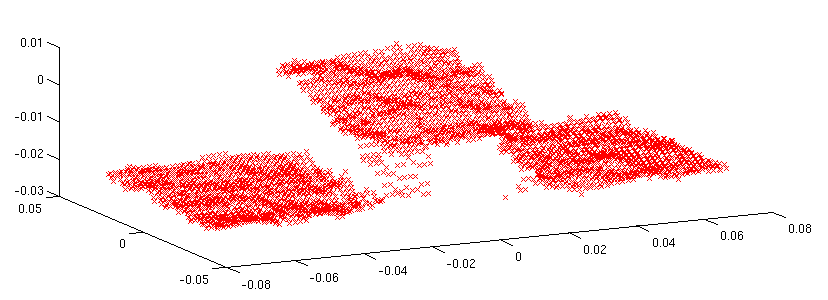}

\caption{Object and its correspondent point cloud and cropped grasping cloud. The top right and bottom images show the volume enclosed by the gripper of an object. The blue points show the selected points of a graspable region of the remote control. The bottom image shows the points enclosed by the gripper volume.}
\label{fig:gripperEnclosingVolumeExample}
\end{figure}

The local shape features are computed in the volume enclosed by the gripper, which is a bounding box located and oriented according to the pre-grasp hypothesis pose. Depth changes in the objects were shown helpful to recognize graspable regions, even in cluttered environments where objects cannot be segmented accurately~\cite{FischingerICRA2013}. The symmetry height accumulated feature~\cite{FischingerICRA2013} is robust, but constrained to top grasps only. We introduce a feature with computations based also on heights, however, it can be computed for any grasping orientation. Our feature, called \emph{depth gradient image}~(DGI), computes the gradient of the depth image in the volume enclosed by the gripper.
%Point of view and plane of projection
This volume defines a depth value (i.e., the height in mm) as the z-component of the distance from the gripper base to the object point. Fig.~\ref{fig:gripperEnclosingVolumeExample} shows an example of the selected region of an object and Fig.~\ref{fig:gripperEnclosingVolume} illustrates the volume of interest enclosed by the gripper.  
%Depth image sampling and gradient computation

%\begin{figure}[t]
% \begin{tabular}{cc}
% %\subfigure[]{\includegraphics[width=.6\columnwidth]{x-yView}}
%\multirow{-7}[]{*}{\subfloat[Gripper and volume of interest]{\includegraphics[width=.45\columnwidth]{gripperAndBB.png}}} & \subfloat[]{\includegraphics[width=.45\columnwidth]{cameraRaw.png}}\\
%&\subfloat[Object point cloud and cropped grasping cloud]
%{
%\includegraphics[width=.245\columnwidth]{pointCloudViewRaw.png}
%\includegraphics[width=.245\columnwidth]{pointCloudCropped.png}
%}
%\end{tabular}
%\caption{Bounding box of the gripper for feature computation. The reference frame of the left side shows the origin for the orthogonal projection of the point cloud onto the height image. The right side images show the volume enclosed by the gripper of an object.}
%\label{fig:gripperEnclosingVolume}
%\end{figure}

\begin{figure}[t]
\centering
\subfloat[Depth Image]{\includegraphics[width=.35\columnwidth]{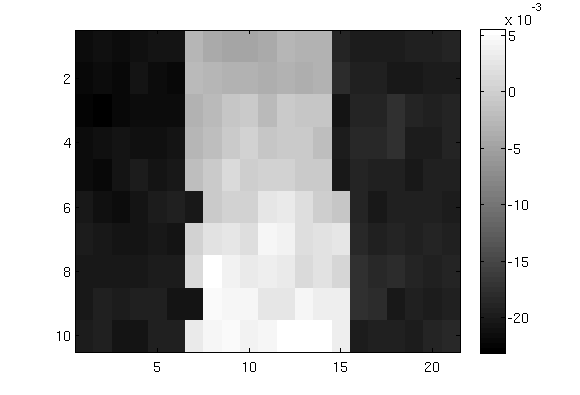}}\subfloat[Depth Gradient Image]{\includegraphics[width=.35\columnwidth]{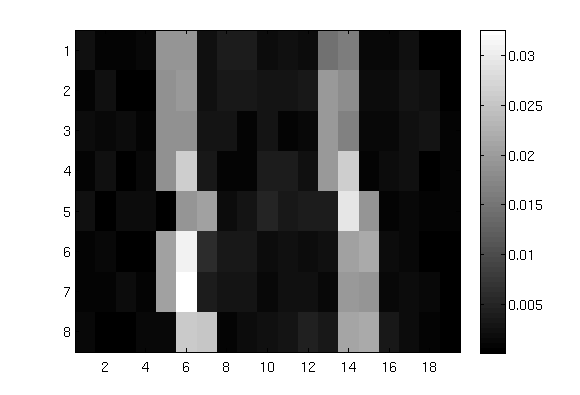}}
\caption{Example of a depth image (10x21 pixels) and its corresponding gradient magnitude (8x19 pixels).}
\label{fig:depthAndDepthGradient}
\end{figure}

The depth image requires a discrete sampling of the volume, which was defined as boxes of 7x7x15 (mm) and is defined as:
\begin{equation}
 \mbox{DI}(u,v)=\left\{ \begin{array}{ll}
       \min_{} \{z\} & \mbox{if $z \in$ box$(u,v)$}\\
       -1 & \mbox{otherwise,}\end{array} \right.
\label{eq:depthImage}
\end{equation}
where box$(u,v)$ represents the set of points inside the box defined at the pixel $(u,v)$. Eq. \eqref{eq:depthImage} performs an orthogonal projection of the closest point to the base of the gripper for every pixel of the depth image. Fig.~\ref{fig:depthAndDepthGradient} shows the depth image for the selected volume in Fig.~\ref{fig:gripperEnclosingVolume}. Finally, the DGI is computed on the depth image by applying pixel differences in $u$ and $v$ as follows:
\begin{eqnarray}
	\mbox{DI}_u(u,v)=\mbox{DI}(u+1,v)-\mbox{DI}(u-1,v),\label{eqn:DIderivative}\\
	\mbox{DI}_v(u,v)=\mbox{DI}(u,v+1)-\mbox{DI}(u,v-1),\label{eqn:DIderivative}\\
	\mbox{DGI}(u,v)=\sqrt{\mbox{DI}_u(u,v)^2+\mbox{DI}_v(u,v)^2}. \label{eqn:depthGradientMagnitude}
\end{eqnarray}
%The main difference is the virtual point of view that is located at the supporting surface of the gripper. 
The DGI acts as a local shape descriptor for grasping prediction. The descriptors of several graspable and non-graspable regions are next fed into a SVM classifier. 

%4. Probability from SVM classifier
\subsection{Grasping probability}
Given DGI shape features $\mathbf x_i$ and their labels $y_i$, we use SVMs~\cite{Cortes1995} with Radial Basis Function (RBF) kernel to discriminate between successful and failed grasps. Before applying the sign function, we map the SVM output:
\begin{align}
f(\mathbf x) = \mathbf w^T \phi (\mathbf x) + b \label{eqn:decisionFunction}\\ 
\phi (\mathbf x_i) = \exp(-\gamma | \mathbf x_i - \mathbf x_j |^2), \label{eqn:rbfKernel}
\end{align}
onto a probability by applying a sigmoid function to the decision value from Eq.~\eqref{eqn:decisionFunction}. We employ the parametric sigmoid:
\begin{align}
P(grasp | local \text{ } shape) = \frac{1}{1+ \exp (A f(\mathbf x) + b)},
\label{eqn:graspingProbability}
\end{align}
where the parameters $A$ and $b$ are obtained by generating a hold-out set and cross-validation. Its advantages were  shown empirically in~\cite{Platt99probabilisticoutputs}.

% The soft margin SVM models better the non-separable training data and reduces the effect of outliers. We select the polynomial kernel of degree two, which performed better than the linear kernel in a set of initial experiments \cite{moreno2011}. 

%5. Dataset and learning result
\subsection{Learning dataset and evaluation}
We train our classifier on the grasping rectangle dataset \cite{JiangMS11}. It contains camera images, point clouds and the pre-grasp poses for both successful and failed grasps. The dataset contains more than 5K samples with similar proportions between positive and negative samples. We remove objects with very noisy point clouds, so in total we have 4708 samples (2424 positive and 2284 negative). We apply cross-validation with $4$ splits in order to find the best RBF parameter ($\gamma = 50$) and sigmoid parameters ($A=2$ and $b=0.5$). The grasping accuracy defined in~\cite{JiangMS11} selects the top grasping region per object and then compares it to the ground truth. 
%make table and improve comments
Table~\ref{tab:graspingAccuracy} shows that our approach using only depth features has good performance, improving the result in~\cite{JiangMS11}. A better result is obtained in~\cite{DBLP:journals/corr/abs13013592}, however they rely on both image and depth features. 

\begin{table}[h]
\centering
\begin{tabular}{|c|c|}
\hline
Approach & Grasping accuracy \\
\hline
\hline
Jiang et. al.  \cite{JiangMS11} &  84.7\\
\hline
Lenz et. al.  \cite{DBLP:journals/corr/abs13013592} &  93.7 \\
\hline
this work &  92.63\\
\hline
\end{tabular}
%\caption{(Perturbed) Pipeline vs. local shape information (PFH) for the ORCA scenarios.}
\caption{Grasping accuracy (\%) on the rectangle dataset~\cite{JiangMS11}.}
\label{tab:graspingAccuracy} 
\end{table}

%%%%%%%%%%%%%%%%%%%%%%%%%%%%%%%%%%%%%%%%%%%%%%%%%%%%%%%%%%%%%%%%%%%%%%%%%%%%%%%%
\section{Experiments}
\label{sec:exp}
%by Laura
We address experimentally the benefits of our proposed pipeline for robot grasping. Specifically, we investigate the following main questions: 

\begin{itemize}
\item[(Q1)] How good is the visual module?
\item[(Q2)] How robust is the probabilistic logic module? How well does it cope with missing information?
\item[(Q3)] Does the integration of high-level reasoning (about task and object category) and low-level learning improve the grasping performance upon local shape features?
\end{itemize}

Since the symbolic visual detection plays an important role for the rest of the pipeline, we first answer question (Q1) by investigating the performance of the object pose and part detection (subsection \ref{sec:pose}). Next, we evaluate the performance of the object category prediction using the manifold-based graph kernel (subsection \ref{sec:object}).   

We answer question (Q2) by tackling in subsections~\ref{sec:object}, \ref{sec:task} and \ref{sec:pregrasp} the following questions:
\begin{itemize}
\item[(i)] does the probabilistic logic module improve upon manifold information for object category prediction?
\item[(ii)] can it predict correctly suitable tasks?
\item[(iii)] can it predict the correct pre-grasp region?
\end{itemize}
We investigate its robustness with respect to object category, task, pre-grasp and grasping point prediction in subsections \ref{sec:object}, \ref{sec:task}, \ref{sec:pregrasp} and \ref{sec:pipeline}. We perturb either the visual observations about the world by dropping the prior on the object category, or the CP-theory by keeping the more general rules.

Finally, to answer question (Q3) we first learn a classifier that maps points sampled from full objects to successful grasps using local shape features. Given a new  object, we then directly predict the most likely grasp using solely local shape information. This is our local shape-based baseline. We compare the baseline with the pipeline classifier, which maps points from semantic pre-grasp object regions to successful grasps, using similar local shape features. Given a new object, we first predict, using high-level reasoning, the most likely object pre-grasp region. We then use the classifier to predict good grasping points among the set of points in the inferred pre-grasp part (the most appropriate for a given task).

\subsection{Datasets and evaluation scenarios}
\label{sec:datasets}
We consider three types of datasets to quantitatively investigate the robustness and power of generalization of our SRL approach. %The two datasets are generated with the help of the ORCA simulator.
For the first type, the object point clouds are obtained from 3D meshes and the object parts are manually labeled. In this case the dataset is synthetic and actual grasps are not executed.  %data acquisition is fully synthetic and the scene description, except the category manifold prior, is provided as ground truth.
For the second type, data samples are obtained from the ORCA simulator~\cite{orcaSimulator}, %For the second dataset the steps of obtaining the scene description are all executed in the ORCA simulator. 
while for the third type, data samples are obtained from a real robot platform.

\subsubsection{Synthetic scenario}

It considers flawless visual detection of objects from 3D meshes. The object points are distributed uniformly on the object surface according to their size by applying the midpoint surface subdivision technique. Point normals are correctly oriented, the object pose and its parts are manually labeled as well as the object containment. This "perfect scenario" serves as an upper-bound comparison scenario to the more realistic scenarios, allowing an extensive evaluation of the generalization capabilities of the probabilistic logic module.
%This scenario assumes that we have the complete point cloud of the input object. The point cloud is obtained from a previously defined 3D mesh of the object by up-sampling points using midpoint surface subdivision technique. The point cloud is imported in the ORCA simulator for scale and point cloud checks. We manually extract from the point cloud the object parts, object pose and containment. Thus, this is a "perfect scenario". On one hand it serves as a comparison baseline for the more realistic setup provided by the second dataset. On the other hand it allows us to evaluate more extensively the power of generalization of the probabilistic logic module across several object categories, while avoiding the problem of extracting object parts for certain objects.
The dataset contains $41$ objects belonging to all categories in our ontology and $102$ grasping scenarios. We denote this dataset $S_{SYN}$.

\subsubsection{ORCA scenarios} 
%KUKA LightWeight Robot (LWR) \cite{KUKAlightweightArm}, the Universal Gripper WSG 50 \cite{universalGripper} with two sensor fingers WSG-FMF \cite{sensorFinger} and an Asus Xtion PRO \cite{AsusXtion}.
The second type of  datasets is used to evaluate all the modules and the full pipeline in simulation. We use ORCA, which provides the sensors (laser range camera Asus Xtion PRO~\cite{AsusXtion} and the Universal Gripper WSG 50~\cite{universalGripper} force sensor), robotic arm (KUKA LightWeight Robot (LWR)~\cite{KUKAlightweightArm}), objects and interface to physics engine (Newton Game Dynamics library~\cite{newtonGD}) for robot grasping simulation. The other modules are external to ORCA and interfaced both with the simulated and real robot. These modules include: object completion, part and pose detection, global shape similarity, probabilistic logical reasoning modules, local shape grasping prediction and the tree-based motion planner~\cite{plannerImpl2003} available in the Open Motion Planning Library (OMPL)~\cite{sucan2012}. Each object is placed on top of a table. 
%We sample the orientation of the object for every discrete pose (several angles upright, several angles upsidedown and several angles sideways) that leads to a static equilibrium of the object. %Different from the fully synthetic scenario, in this case, we use a more realistic point cloud of the input object. The point cloud is simulated laser range data and is estimated
%from several view points using object symmetries. The 3D data for each view is acquired from a simulated range camera (with the parameters of the Kinect sensor), placed on the robot platform~\ref{figure_to_be_added}. For each object we use between one and eight views depending on whether the object is symmetric or not. The steps of obtaining the scene description are executed in the ORCA simulator.

Further, we consider four possible settings. In the first setting, object pose is not estimated but given by the ground truth, while the parts are estimated from the completed point cloud, as explained in Section~\ref{sec:module1}. Thus, the scene description may have missing parts when they are occluded or not detected. Additionally, we assign to all detected parts probability $1.0$. We denote this dataset $S_{REAL\_semi}$. In the second setting, both object pose and its parts are estimated from the completed point cloud. While the pose has associated a likelihood, we keep highly confident parts. We denote this setting $S_{REAL}$. In the third setting we also provide a part likelihood according to the limitations of the detection algorithm. We denote this dataset $S_{REAL\_noisy}$. Finally, the fourth setting includes actual grasping tests with the simulated robot. It comprises a subset of the scenarios from the third setting, where all containers are empty and all 
objects are graspable by the robot. The rationale behind is that it is very difficult to check whether a container is full or if some objects (too big or too small) do not fit the gripper capabilities.  We denote this dataset $S_{GRASP\_noisy}$. In addition, object poses considered are \emph{upright} or \emph{sideways} due to the ambiguity between upright and upside-down when using global shape representations. Each of the first three settings contains 26 different objects, instances of categories \emph{pan, bowl, cup, glass, bottle, can, hammer, screwdriver, knife} and $126$ grasping scenarios. The fourth setting contains 18 objects, instances of categories \emph{pan, cup, glass, bottle, can, hammer, screwdriver, knife} and $113$ grasping scenarios.%We consider different view points of the camera, which, applied on the same objects and similar poses, render different scenarios.

\begin{figure}[t]
\includegraphics[width=0.16\textwidth]{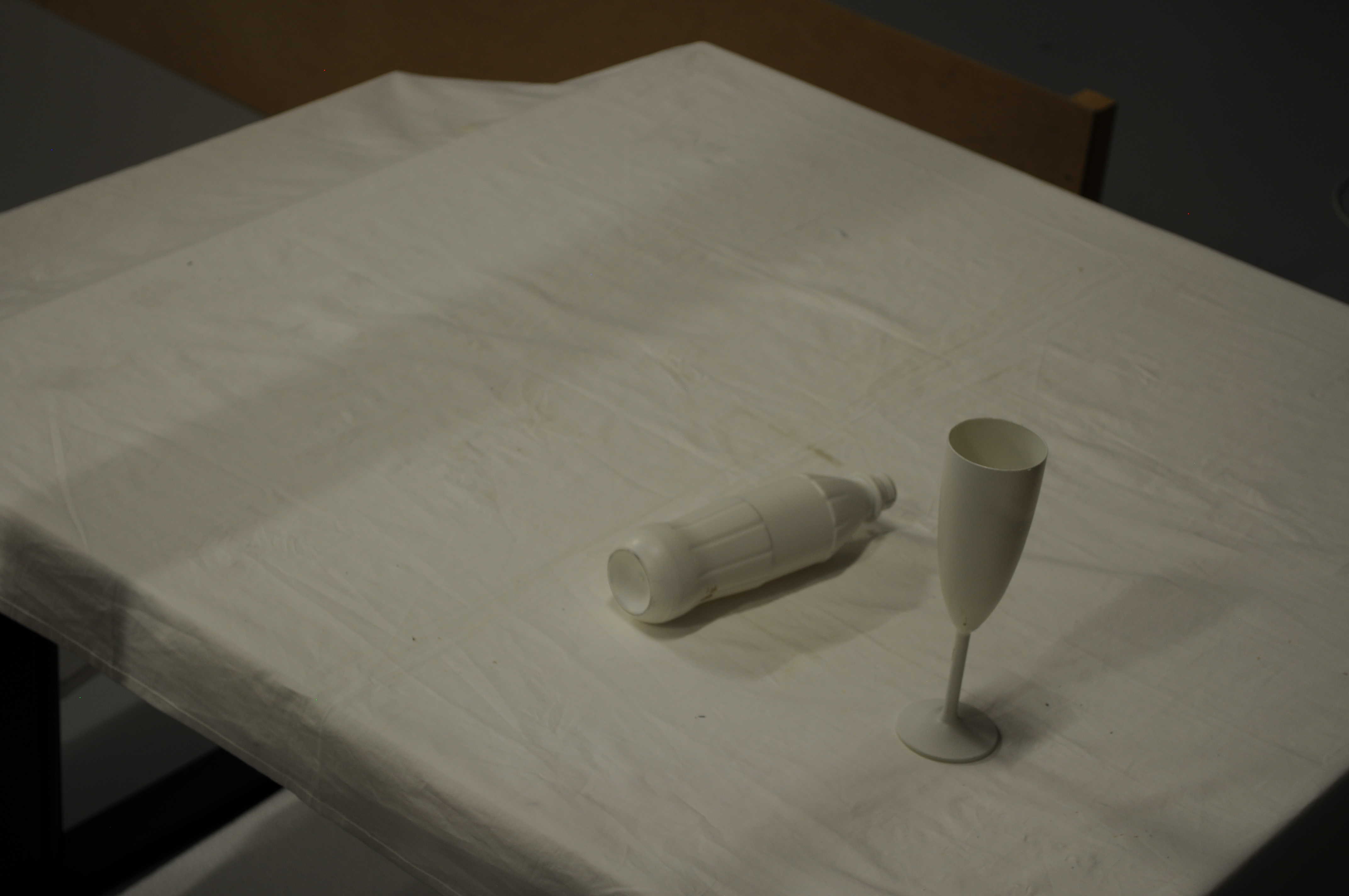}\includegraphics[width=0.16\textwidth]{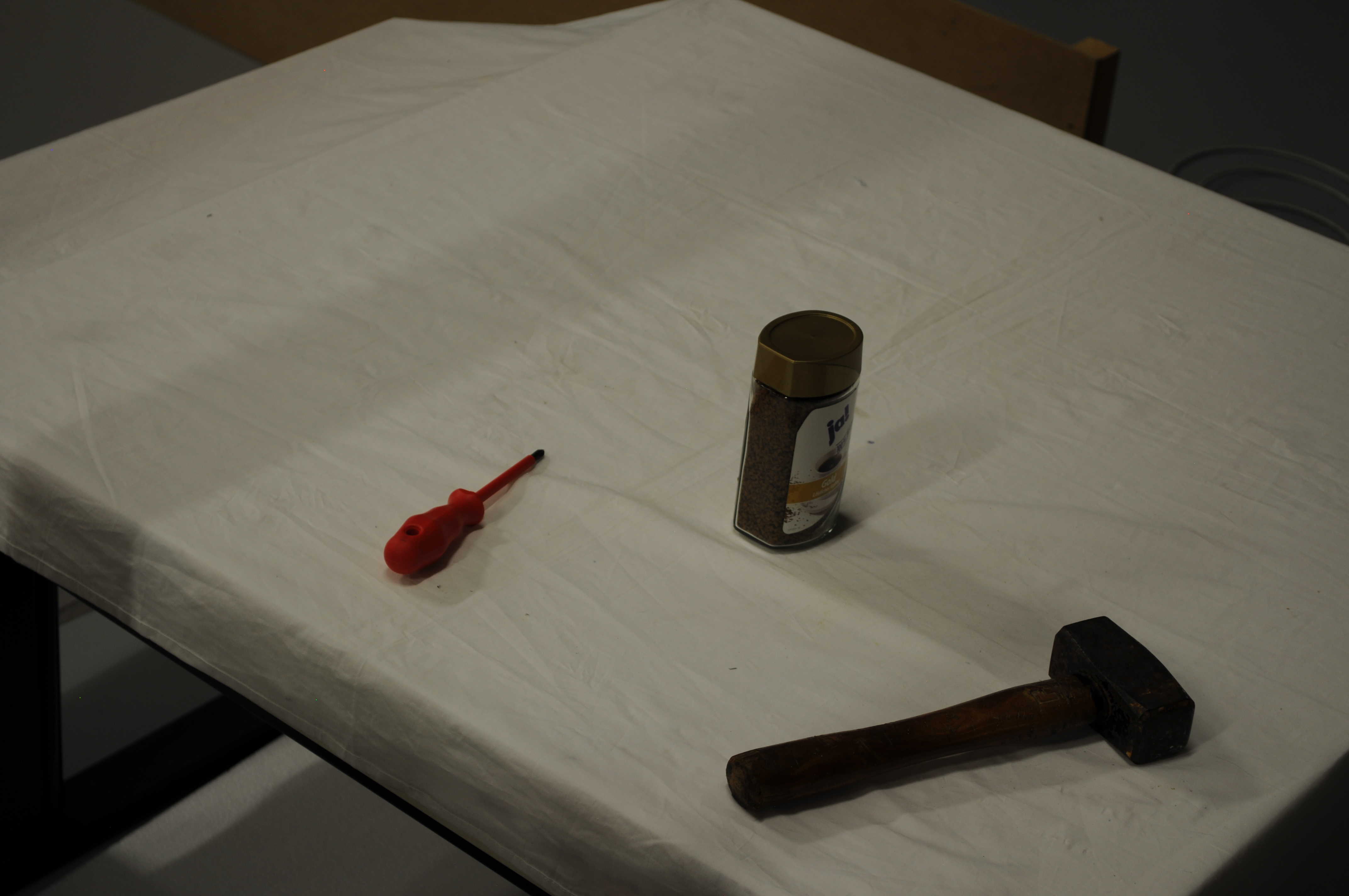}\includegraphics[width=0.16\textwidth]{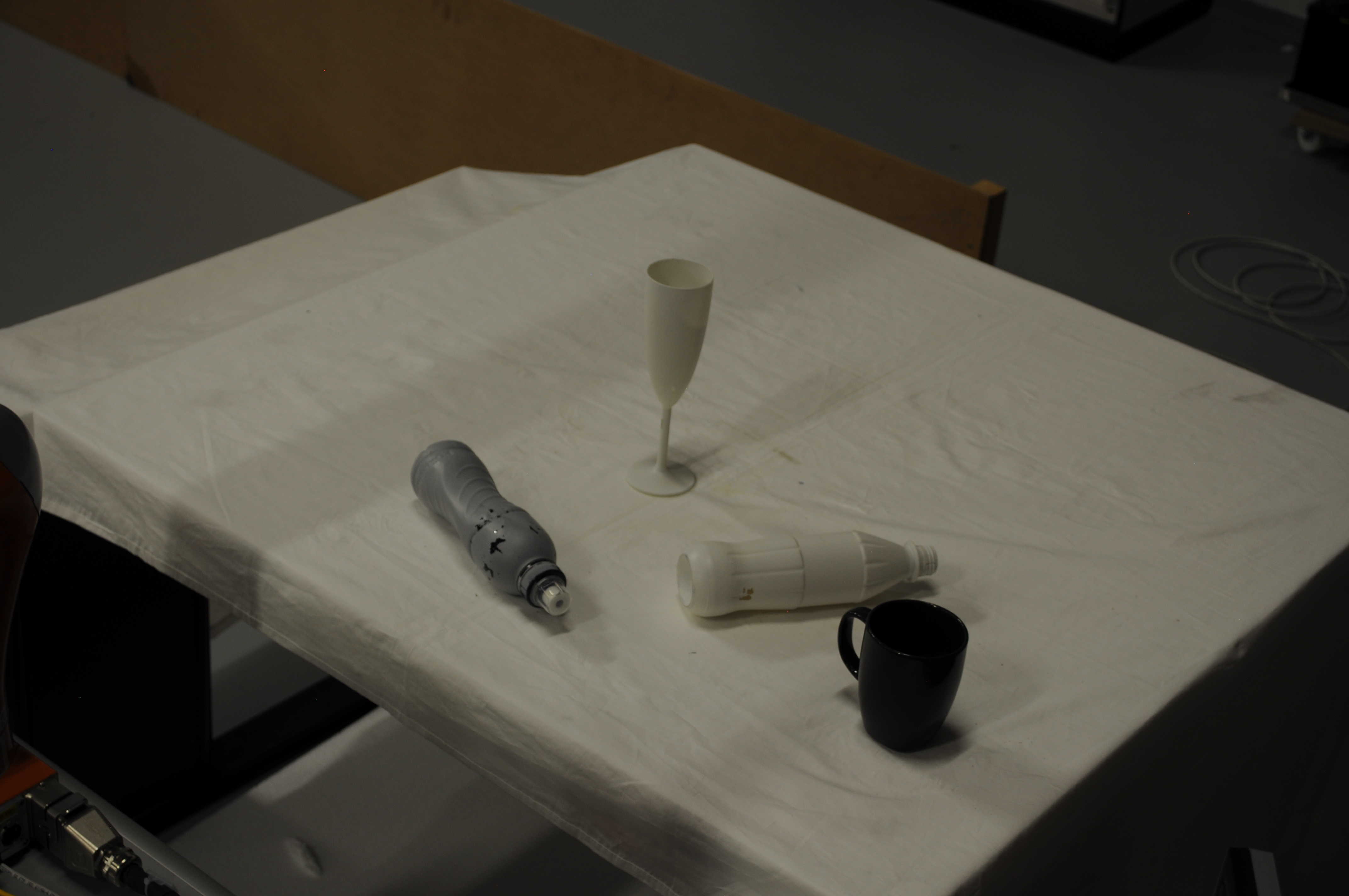}
\caption{Experimental settings with the real robot. Each picture shows the objects utilized for each scenario. Additional object constraints are: the gray bottle of scenario3 is full with water, the white bottle is empty and the coffee container is full of coffee.}
\label{fig:realRobotScenarios} 
\end{figure}

\subsubsection{Real robot scenario}
\begin{figure}
\includegraphics[width=1\columnwidth]{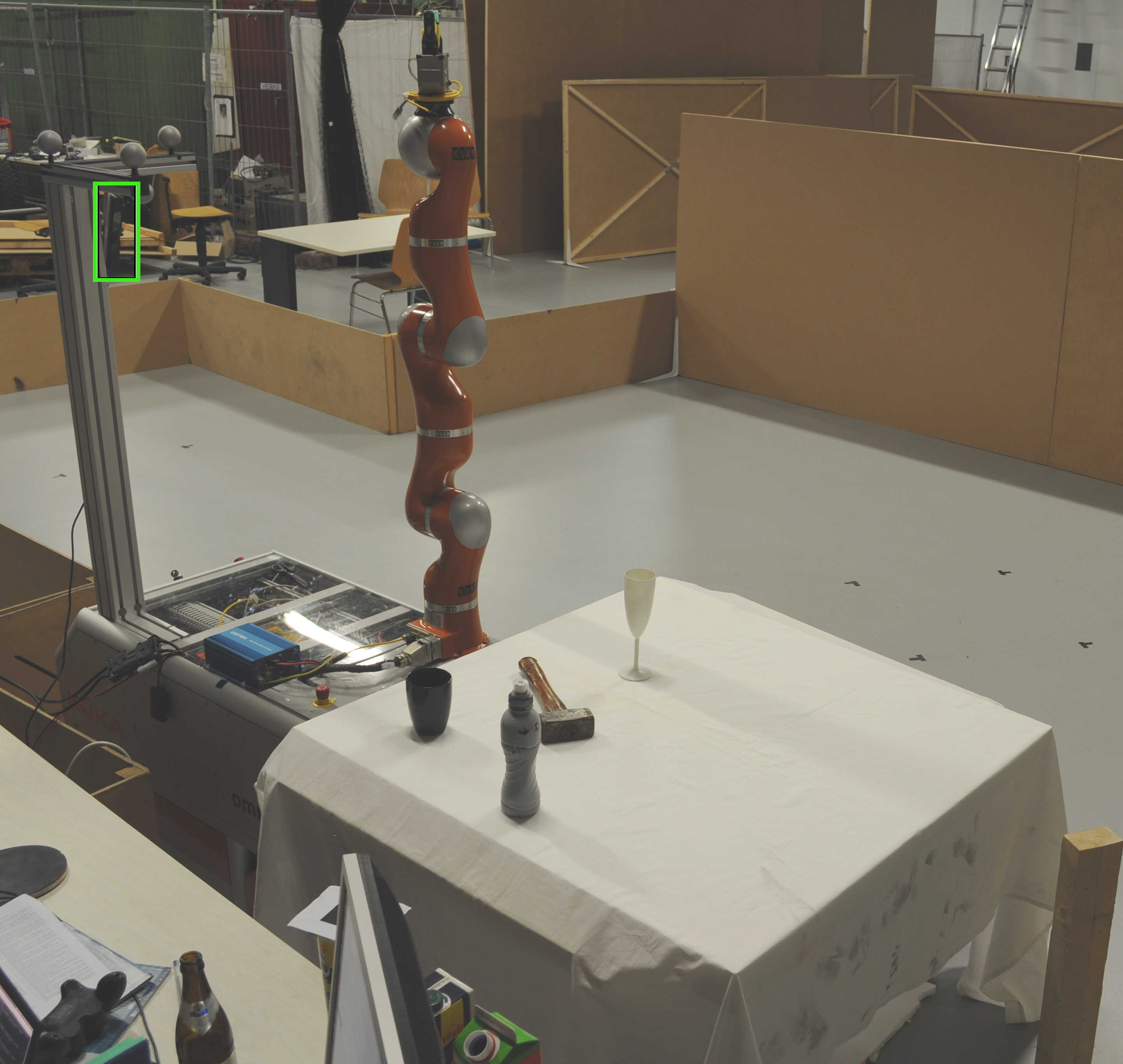}
\caption{The table is in front of the mobile platform. The range sensor is marked by the green rectangle.}
\label{fig:experimentalSetup}
\end{figure}

The robot scenario considers the same type of tests as those included in the $S_{GRASP\_noisy}$. Fig.~\ref{fig:experimentalSetup} shows our setup. In addition, we evaluate the performance of the pipeline when two or more objects are in the field of view of the camera and the field of action of the arm, considering three settings of increasing complexity in terms of path planning. The less complex setting (scenario1), considers only two objects which are instances of \emph{glass}  and \emph{bottle}, in a way that planning constraints are very similar to a single object on the table. The setting with intermediate complexity (scenario2) includes three objects which are instances of \emph{can, hammer} and \emph{screwdriver}. The more complex setting (scenario3) considers four objects which are instances of \emph{bottle, glass} and \emph{cup}. Fig.~\ref{fig:realRobotScenarios} shows the objects of every scenario. In addition to the larger number of objects, we also consider object placement as another criterion 
for evaluation. Object placement is performed in two steps: (i) plan from the grasp pose to a post-grasp pose and (ii) plan from the post-grasp pose to the grasp pose. We denote the associated dataset $S_{ROBOT}$.

The synthetic dataset and part of ORCA datasets are available for download at \url{http://www.first-mm.eu/data.html}.

\subsection{Evaluation measures}

We evaluate our experiments in terms of accuracy given by $\frac{\# successes}{\# tries} \cdot 100\%$. We assess a $success$ in several ways. Depending on the prediction task, the ground truth is either one value (object categorization, pose detection) or a set of values (task and pre-grasp prediction, part detection). 

In the case of object part detection we consider a success if all detected object parts match the ground truth. A correct pose detection is considered when the discrete pose predicted matches the ground truth.

For object categorization we take as prediction the category with the highest probability and consider it a success if it matches the ground truth category. For the uniform prior, it can be that two or more categories are predicted with the same probability. This case is reported as a false positive.

%In this case we evaluate a success in two ways. We denote with $E^{cat}_1$ the evaluation setting in which all the (equally) maximum scored categories are equal with the ground truth category, and with $E^{cat}_2$ the setting in which one of the (equally) maximum predictions is equal to the ground truth. The two settings are the same when exactly one maximum is predicted (the manifold prior case). 

For task/pre-grasp prediction evaluation, the ground truth $Gt$ of each instance is a set (e.g., the tasks \emph{p\&pOn} and \emph{p\&pInUpright} may be equally possible in a particular scenario). In this case, we compare the set of best predictions $Pr$ to $Gt$, where $|Pr|\leq|Gt|$. If $Pr\subseteq Gt$ a success is reported. We present results for different sizes of $Pr$, such that $|Pr|$ belongs to the set \{$|Gt|-i$\}, with $i$ ranging from $0$ (the most restrictive evaluation setting) to $|Gt|-1$ (the most pertinent setting). We denote the possible evaluation settings as $E_i$. For the scenarios with robot grasping execution ($S_{GRASP\_noisy}$ and $S_{ROBOT}$) the evaluation must consider the success of grasping execution with respect to the valid grasping hypotheses provided by the pipeline. In this case the accuracy is given by $\frac{\# correctly \text{ } estimated \text{ } task/part }{\# valid \text{ } grasping \text{ } hypotheses} \cdot 100\% $.

\subsection{Results and discussion} 

In the following we present quantitative experimental results for all questions. For all results bold font, if present, indicates the best performance.

\subsubsection{Object pose and part detection}
\label{sec:pose}

\begin{table}[t]
\centering
\begin{tabular}{|c|c|c|}
\hline
Dataset & Part detection & Pose detection \\
\hline
\hline
%\multirow{3}{*}{$S_{REAL\_noisy}}$   &  gt 50\%  &  63.33 \%& 35.40\% & 42.72\%\\
$S_{GRASP\_noisy}$ & 84.56 & 100\\
\hline
$S_{ROBOT}$ &  82.14 & 100 \\
\hline
\end{tabular}
%\caption{(Perturbed) Pipeline vs. local shape information (PFH) for the ORCA scenarios.}
\caption{Accuracy (\%) for object part and pose detection.}
\label{tab:visualRepresentation} 
\end{table}

Results are shown in Table~\ref{tab:visualRepresentation}. We note that the PCA global representation is able to cope well with object pose detection, considering the table-top assumption and the object categories assumed. We note that we do not consider the upside-down pose for these tests, as in real-world applications usual poses are upright and sideways. Object part detection suffers from part occlusion for particular object poses, reducing the pipeline performance for object category prediction. In addition to the accuracy, we stress the execution time for the object completion using symmetries and pose detection. The average execution times are 27.5 ms and 15.71 ms on a PC using one core of the Intel Xeon (2.67GHz). These numbers confirm the computational efficiency of our approach, which allows to make fast decisions.

\subsubsection{Object category prediction}
\label{sec:object}
%We evaluate object category prediction for all datasets for the manifold-based graph kernel in %Table~\ref{tab:object_category_graphkernel} and for the probabilistic logic module (PLM) in Table~\ref{tab:object_category_plm}.

% \begin{table}[ht]
% \centering
% \begin{tabular}{|c|c|c|c|c|}
% \hline
% $ $ Dataset $ $ &  $ $ Measure $ $  &  $ $ Random $ $  &  $ $ Label fractions $ $  &  $ $ Manifolds $ $  \\
% \hline
% \hline
% \multirow{2}{*}{$S_{SYN}$}  	&  $acc$		&9.1\%  &63.4\%  	& \textbf{87.8\%}\\
% 				&  $rank$		&5.5  	&1.88  		& \textbf{1.39} \\
% \hline
% \multirow{2}{*}{$S_{REAL}$}  	& $acc$			&9.1\%  &\textbf{39.7}\%  	&\textbf{39.7\%} \\
% 				& $rank$		&5.5  	&2.54  		&\textbf{2.45} \\
% \hline
% \multirow{2}{*}{$S_{GRASP\_noisy}$}  	& $acc$	&  9.1 \% & - 	&\textbf{37.95\%} \\
% 				& $rank$		& 5.5 	&-  		&\textbf{3.95} \\
% \hline
% \multirow{2}{*}{$S_{ROBOT}$}  	& $acc$ &	 9.1 \%& -  	&\textbf{50\%} \\
% 				& $rank$		& 5.5 	& -		&\textbf{3.1} \\
% \hline
% \end{tabular}
%   \caption{ Accuracy ($acc$, the higher the better) and average rank of the true category ($rank$, the lower the better): propagation kernel (Manifolds) vs. baselines for object categorization.}
% \label{tab:object_category_graphkernel} 
% \end{table}

\begin{table*}
\centering
\begin{tabular}{|c|c|c|c|c|c|c|c|}
\hline
$ $ Dataset $ $ &  $ $ Random $ $  &  $ $ Label fractions $ $  &  $ $ Manifolds $ $ & PLM$_{uniform}$ & PLM$^{general}_{uniform}$ & PLM  & PLM$^{general}$ \\
\hline
\hline
\multirow{1}{*}{$S_{SYN}$}  	& 9.1  & 63.4  	& {\it 87.8} & 31.37 & 31.37 & {\bf 93.14}  &  92.16 \\
%				&  $Rank$		&5.5  	&1.88  		& \textbf{1.39} & & & & \\
\hline
\multirow{1}{*}{$S_{REAL\_semi}$}  	& 9.1  & {\it 39.7}  	& {\it 39.7} & 14.29 & 14.29 & {\bf 49.21} & {\it 46.83} \\
%				& $Rank$		&5.5  	&2.54  		&\textbf{2.45} & & & & \\
\hline
\multirow{1}{*}{$S_{REAL}$}  & 9.1  & {\it 39.7}  	& {\it 39.7} & 14.29 & 14.29 & {\bf 48.41} & {\it 46.83}\\
%				& $Rank$		&5.5  	&2.54  		&\textbf{2.45} & & & & \\
\hline
\multirow{1}{*}{$S_{REAL\_noisy}$}  	& 9.1  & {\bf 39.7}  	& {\bf 39.7} & 14.29 & 14.29 & {\bf 39.7} & {\bf 39.7}\\
%				& $Rank$		&5.5  	&2.54  		&\textbf{2.45} & & & & \\
\hline
\multirow{1}{*}{$S_{GRASP\_noisy}$}  	& 9.1  & - 	& \textbf{37.95} & - & -& -& -\\
%				& $Rank$		& 5.5 	&-  		&\textbf{3.95} & & & & \\
\hline
\multirow{1}{*}{$S_{ROBOT}$}  	& 9.1 & -  	& \textbf{50} & -& -& -& -\\
%				& $Rank$		& 5.5 	& -		&\textbf{3.1} & & & & \\
\hline
\end{tabular}
  \caption{ Accuracy (\%): PLM vs. propagation kernel (Manifolds) vs. baselines for object categorization.}
\label{tab:object_category_graphkernel} 
\end{table*}

We next evaluate the object category prediction task and report accuracy results in Table~\ref{tab:object_category_graphkernel}. We compare random category assignment (Random), a linear kernel among label counts (Label fractions), propagation kernels (Manifolds) and the probabilistic logic module (PLM). Note, that the second baseline method corresponds to setting $T=0$ in Eq.~\eqref{equ:prop_kernel}. Hence, we compute a linear base kernel using only label counts as features and no manifold information (graph structure) is considered. 

For Manifolds we set the parameter $T$ to give the best leave-one-out accuracy performance $T=12$ on $S_{SYN}$ and $T=0$ on all $S_{REAL}$.\footnote{The Manifolds results for $S_{REAL\_semi}$, $S_{REAL}$ and $S_{REAL\_noisy}$ are the same due to the fact that object pose and part confidence are not used by the propagation kernel. We refer to these datasets as $S_{REAL}$ for the Manifolds evaluation.} On $S_{SYN}$ we clearly see that incorporating manifold information, that is, using the graph kernel, improves the accuracy more than $20\%$ upon using label information only. On $S_{REAL}$ the manifold improvement is $4$ times more than random.

We also evaluated the Manifolds and baseline approaches in terms of average rank ($r$) of the true category. We obtain $r = 1.39$ ($S_{SYN}$) and $r = 2.45$ ($S_{REAL}$) for Manifolds, $r = 1.88$ ($S_{SYN}$) and $r = 2.54$ ($S_{REAL}$) for Label fractions and $r = 5.5$ ($S_{SYN}$, $S_{REAL}$) for Random. Note that for the rank lower values are better and the optimal result would be $1$, i.e., the true category has the highest probability for all objects in the respective dataset/scenario. On both $S_{SYN}$ and $S_{REAL}$ propagation kernels decrease the average rank compared to Label fractions and Random.  The parameter $T$ for Manifolds was set in this case to $T=13$ on $S_{SYN}$ and $T=3$ on $S_{REAL}$. The main problem for predicting the object category in the $S_{REAL}$ scenarios is that the part assignment is achieved by applying the part detector introduced in Section~\ref{sec:partdetection}. As this detector was designed to work for general objects (i.e.\ we assume the scenarios to be as realistic as possible) its performance is rather poor for some object categories. Overall, manifold information leads to a good prior distribution among object categories (more details can be found in~\cite{Neumann13mlg}).

Next, we incorporated the manifold priors into the PLM. We started from the priors that gave the best accuracy for each task. For object categorization this translates into $T=13$ for $S_{SYN}$ and $T=0$ for $S_{REAL}$. To show the robustness of the PLM, we varied the generality of our categorization theory and experimented with and without the manifold prior on the object category. $PLM^{general}_{uniform}$ indicates the more general theory setting with the uniform prior, while  $PLM^{general}$ indicates the more general theory with the manifold prior. The results show that the PLM improves object categorization accuracy upon manifold information. By increasing the generality of the theory, we do not loose much in terms of performance when the manifold information is used, and we are still able to improve upon the prior.

Note that by removing the manifold prior, the PLM still gives a reasonable result ($3$ times better than Random on $S_{SYN}$). We also evaluated the PLM$_{uniform}$ with a second accuracy definition ($acc$) in which a $success$ is reported if at least one category in the set of equally and maximum category prediction values is equal to the ground truth. In this case we obtain $acc = 68.25\%$ ($S_{REAL}$ datasets) and $acc = 99.02\%$ ($S_{SYN}$) for PLM$_{uniform}$, and $acc = 62.70\%$ ($S_{REAL}$ datasets) and $acc = 94.12\%$ ($S_{SYN}$) for $PLM^{general}_{uniform}$. The results obtained using this evaluation setting explain the good performance for the other grasping tasks in subsections~\ref{sec:task} and~\ref{sec:pregrasp}. That is, estimating the category of an object as any of the sub-categories of a super-category in the ontology is satisfactory to predict good semantic pre-grasps.

% \begin{table}[ht]
% \centering
% \begin{tabular}{|c|c|c|c|c|c|c|c|}
% \hline
% Dataset & Acc & PLM$_{uniform}$ & PLM$^{general}_{uniform}$ & PLM  & PLM$^{general}$\\
% \hline
% \hline
% \multirow{2}{*}{$S_{SYN}$} & $E^{cat}_1$ & 31.37 & 19.61 & {\bf 96.08}  &  93.14 \\
%  & $E^{cat}_2$ & {\bf 99.02} & 70.59 &  96.08 & 93.14 \\
% \hline
% \multirow{2}{*}{$S_{REAL\_semi}$}& $E^{cat}_1$ & 14.29 & 14.29 & {\bf 46.83} & 39.7 \\
%  & $E^{cat}_2$ &  {\bf 68.25} & 62.70 & 46.83 & 53.17\\
% \hline
% \multirow{2}{*}{$S_{REAL}$}& $E^{cat}_1$ & 14.29 & 14.29 & {\bf 46.83} & 39.7 \\
%  & $E^{cat}_2$ & {\bf 68.25} & 62.70 & 46.83 & 39.7\\
% \hline
% \multirow{2}{*}{$S_{REAL\_noisy}$}& $E^{cat}_1$ & 14.29 & 14.29 & {\bf 39.7} & {\bf 39.7} \\
%  & $E^{cat}_2$ & {\bf 68.25} & 62.70 & 39.7 & 39.7\\
% \hline
% \end{tabular}
% \caption{Accuracy (\%): PLM for object categorization.}
% \label{tab:object_category_plm} 
% \end{table}

% \begin{table}[ht]
% \centering
% \begin{tabular}{|c|c|c|c|c|c|c|c|}
% \hline
% Dataset & PLM$_{uniform}$ & PLM$^{general}_{uniform}$ & PLM  & PLM$^{general}$\\
% \hline
% \hline
% \multirow{1}{*}{$S_{SYN}$} & 31.37 & 19.61 & {\bf 96.08}  &  93.14 \\
% \hline
% \multirow{1}{*}{$S_{REAL\_semi}$} &  14.29 & 14.29 & {\bf 46.83} & 39.7 \\
% \hline
% \multirow{1}{*}{$S_{REAL}$} & 14.29 & 14.29 & {\bf 46.83} & 39.7 \\
% \hline
% \multirow{1}{*}{$S_{REAL\_noisy}$} & 14.29 & 14.29 & {\bf 39.7} & {\bf 39.7} \\
% \hline
% \end{tabular}
% \caption{Accuracy (\%): PLM for object categorization.}
% \label{tab:object_category_plm} 
% \end{table}

\subsubsection{Task prediction}
\label{sec:task}

\begin{figure}[t]
 \begin{center}
  \includegraphics[width=6cm, height=8cm]{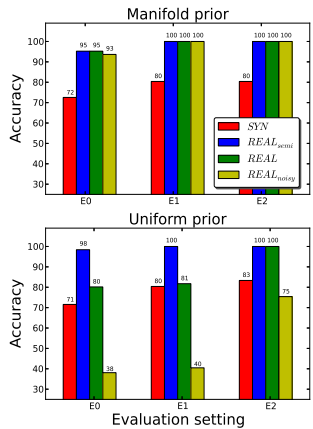}
 \end{center}
\caption{Accuracy (\%): PLM for task prediction using all evaluation settings.}
\label{fig:task_plm} 
\end{figure}

We answer the question \emph{what is the suitable task} by reporting results in Table~\ref{tab:task_plm_e0} (top rows) with evaluation setting $E_0$ for both general and more specific object categorization theories. Fig.~\ref{fig:task_plm} presents results for the specific object categorization theory using all evaluation settings. We experimented on all datasets with and without the manifold prior. The presence of the prior gives significantly better results in the most restrictive evaluation setting. For the other evaluation settings, in most situations, the probabilistic logic module will return, although not as the first option, a correct possible task with or without a prior. 

In the scenarios with grasp execution (bottom rows in Table~\ref{tab:task_plm_e0}), the evaluation settings $E_0$ and $E_1$ consider the outcome of the grasping action. The additional source of failures on grasping include uncertainty on the pose of the objects and the gripper, which are caused by the sensor and the object completion. These sources have effects on the performance of the planner, for instance placing the gripper a bit misaligned or hitting and object before closing the gripper. The few cases where the uniform prior provided better results ($S_{ROBOT}$) are explained by the fact that the complete pipeline failed on the PLM experiment more than the uniform prior due to uncertainty.

We note the importance of having a prior probability distribution over the object categories, rather than the top category. We perform the same experiments only with the top predicted category and obtain accuracies of $95.24\%$, $89.68\%$ and $89.68\%$ for $S_{REAL\_semi}$, $S_{REAL}$ and $S_{REAL\_noisy}$, respectively, which are lower than using the full prior (see Table~\ref{tab:task_plm_e0}).

%Scenarios $S_{ROBOT}$ and $S_{GRASP\_noisy}$ in Table \ref{tab:task_plm_e0} have two different accuracies, where the left side one is given by $Acc = \frac{\# correctly \text{ } estimated \text{ } tasks}{\# successful \text{ } grasps}$ and the right side one is $(Acc) = \frac{\# correctly \text{ } estimated \text{ } tasks}{\# valid \text{ } grasping \text{ } hypotheses}$. The right side accuracy takes into account the grasping failures due to perception and planning errors, for instance placing the gripper a bit misaligned or hitting and object before closing the gripper.

\subsubsection{Pre-grasp selection}
\label{sec:pregrasp}
To answer the question \emph{what is the suitable pre-grasp} we considered the specific object categorization theory and experimented with both a more specific and a more general task-dependent grasping theory. The results for both settings when the task is given for evaluation setting $E_0$ are shown in Table~\ref{tab:pregrasp_taskgiven_e0}. We note that increasing the generality of the model does not cause much performance loss. The result confirms generalization over similar object parts and object/task categories, which implies that if the input object is an unseen category, such as a paint roller or a vase the grasping pipeline is robust enough to return a good grasping part. This allows us to experiment with a wide range of object/task categories and lets us to believe that our approach can be extended beyond the categories used, by augmenting the probabilistic logic module with extra rules. 

\begin{figure}[t]
 \begin{center}
  \includegraphics[width=6cm, height=8cm]{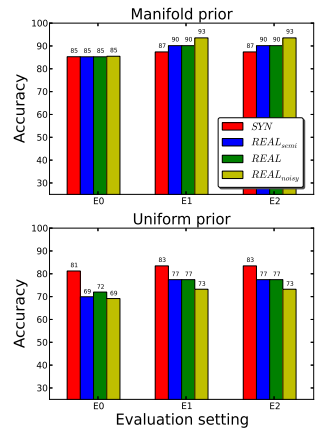}
 \end{center}
\caption{Accuracy (\%): PLM for pre-grasp prediction for all evaluation settings.}
\label{fig:pregrasp_taskgiven_plm} 
\end{figure}

When the task is  predicted by the probabilistic logic module we report accuracies of $95.10\%$ and $100.0\%$ for $S_{SYN}$ and $S_{REAL\_noisy}$, respectively, for both priors using the evaluation setting $E_0$. When the task is given, results for the pre-grasp prediction task using the most specific task-dependent grasping theory for all evaluation settings are reported in Fig.~\ref{fig:pregrasp_taskgiven_plm}. 

\begin{table}[t]
\centering
\begin{tabular}{|c|c|c|c|c|c|c|}
\hline
Dataset & PLM$_{uniform}$ & PLM$^{general}_{uniform}$ & PLM & PLM$^{general}$\\
\hline
\hline
$S_{SYN}$ & 71.57 & 65.69 & {\bf 72.55} & {\bf 72.55} \\
$S_{REAL\_semi}$ & {\bf 98.41} & {\bf 98.41}  & 95.24 & 93.65 \\
$S_{REAL}$  & 80.16 & 80.16 & {\bf 95.24} & 93.65 \\
$S_{REAL\_noisy}$ & 38.10 & 38.10 & {\bf 93.65} & {\bf 93.65} \\
\hline
$S_{GRASP\_noisy}$ $E_0$& 30.00 & - & {\bf 35.71} & - \\
$S_{GRASP\_noisy}$ $E_1$& 40.00 & - & {\bf 50.00}& - \\
$S_{ROBOT}$ $E_0$& {\bf 28.57} & - & 25.00& - \\
$S_{ROBOT}$ $E_1$& {\bf 85.71} & - & 75.00 & - \\
\hline
\end{tabular}
\caption{Accuracy (\%): PLM for task prediction. }
\label{tab:task_plm_e0} 
\end{table}

\begin{table}[t]
\centering
\begin{tabular}{|c|c|c|c|c|}
\hline
Dataset & PLM$_{uniform}$ & PLM$^{general}_{uniform}$ & PLM & PLM$^{general}$\\
\hline
\hline
$S_{SYN}$ & 81.23 &  80.67 & {\bf 85.29} & 84.73\\
$S_{REAL\_semi}$  & 69.95 & 72.00 & 85.26 & {\bf 86.73} \\
$S_{REAL}$  & 72.00 & 72.00 & {\bf 85.26} & 84.69 \\
$S_{REAL\_noisy}$ & 69.16 & 69.16 & 85.49 &  {\bf 86.73} \\
\hline
$S_{GRASP\_noisy}$ $E_0$& 66.7 & - & {\bf 75.51} & - \\
$S_{GRASP\_noisy}$ $E_1$& 66.7 & - & {\bf 75.51} & - \\
$S_{ROBOT}$ $E_0$& {\bf 66.7 } & - & {\bf 66.7} & - \\
$S_{ROBOT}$ $E_1$& {\bf 66.7} & - & {\bf 66.7} & - \\
\hline
\end{tabular}
\caption{Accuracy (\%): PLM for pre-grasp prediction. }
\label{tab:pregrasp_taskgiven_e0} 
\end{table}

For this task as well, in the scenarios with grasp execution, the evaluation considers the outcome of the grasping action. In all cases the object category provided by the manifold matching has better performance than the uniform category prior, showing the benefits of the pipeline. Again, it is important to consider the full prior distribution as input to the PLM, instead of only the top category. Our experiments using solely the top category resulted in accuracies of $81.63$, $81.63$ and $82.88$, respectively, for the three $S_{REAL}$ datasets, which are lower than the ones using the full prior. For both task and pre-grasp prediction we incorporated the manifold priors using parameters $T=13$ for $S_{SYN}$ and $T=3$ for $S_{REAL}$.

The robot experiments ($S_{GRASP\_noisy}$ and $S_{ROBOT}$) did not consider the mutually exclusiveness assumption. However, for the other datasets we performed experiments both with and without this assumption. The means of the differences between the mutually exclusive and the non-exclusive results for object categorization, task prediction and pre-grasp prediction ($1.6\%$, $8.5\%$ and $2.6\%$, respectively) indicate a better result for the mutually exclusive setup. This lets us to believe that current robot results can be only improved by the exclusiveness assumption.

\subsubsection{Full pipeline evaluation}
\label{sec:pipeline}
The selection of grasping points using only local shape descriptors bias the ranking of the object points towards the most "visually graspable", disregarding other constraints such as the pre-grasp pose for task execution, path planning and post-grasp object pose. By adding those constraints, regions with lower "visually graspable" probability will become more important when considering the task execution probability and vice versa. Thus, in order to answer question (Q3) we compute the percentage of grasping points that have a low "visually graspable" probability from Eq.~\eqref{eqn:graspingProbability}, but still lead to successful grasps when taking into account task constraints. Results are shown in Table~\ref{tab:pipeline}, having as baseline the local shape-based approach. We also investigate its robustness with respect to grasping point prediction by dropping the prior on the object category (Pipeline$_{uniform}$).

\begin{table}[t]
\centering
\begin{tabular}{|c|c|c|c|c|}
\hline
Feature (DGI) & Measure & Local & Pipeline & Pipeline$_{uniform}$ \\
\hline
\hline
%\multirow{3}{*}{$S_{REAL\_noisy}}$   &  gt 50\%  &  63.33 \%& 35.40\% & 42.72\%\\
\multirow{3}{*}{$S_{GRASP\_noisy}$}   &  lt 0.5  &  44.55\% & \textbf{52.38\%} & 50\%\\
%&  gt 40\%  &  80 \%& 68.14\% & 74.76\% \\
&  lt 0.4  &  27.27\% & \textbf{28.57\%} & 21.15\% \\
%& gt 30\% &  96.67 \%& 88.5\% & 91.26\%  \\
& lt 0.3 &  4.55\% & \textbf{7.94\%} & 7.69\%  \\
\hline
\multirow{3}{*}{$S_{ROBOT}$} &  lt 0.5  &  37.5\% & \textbf{50\%} & 46.15\% \\
&  lt 0.4  &  25\% & \textbf{42.86\%} & 23.08\% \\
&  lt 0.3  &  12.5\% & \textbf{28.57\%} & 23.08\% \\
\hline
\end{tabular}
%\caption{(Perturbed) Pipeline vs. local shape information (PFH) for the ORCA scenarios.}
\caption{Percentage (\%) of successfully graspable points that have "visually graspable" probability less than (lt) 0.3, 0.4 or 0.5: Pipeline vs. local shape grasp prediction.}
\label{tab:pipeline} 
\end{table}

The full pipeline selects in average more points having low "visually graspable" probabilities than the other options. This behavior confirms the importance of task constraints on the computation of the grasping probability. We note that  the complete pipeline clearly improves upon using the local shape-based approach. This answers affirmatively (Q3). Additionally, the results obtained by Pipeline$_{uniform}$ show again the robustness of the pipeline.

% \begin{table}[ht]
% \centering
% \begin{tabular}{|c|c|c|}
% \hline
% Dataset & PLM$_{uniform}$ &  PLM \\
% \hline
% \hline
% $S_{GRASP\_noisy}$ $E_0$& 30 & {\bf35.71}  \\
% $S_{GRASP\_noisy}$ $E_1$& 40 &{\bf 50} \\
% $S_{ROBOT}$ $E_0$& {\bf 28.57} & 25 \\
% $S_{ROBOT}$ $E_1$& {\bf 85.71} & 75 \\
% \hline
% \end{tabular}
% \caption{Pipeline accuracy $acc_{p}$ (\%): Task prediction.}
% \label{tab:pipelineAccuracyTask} 
% \end{table}

% \begin{table}[ht]
% \centering
% \begin{tabular}{|c|c|c|}
% \hline
% Dataset & PLM$_{uniform}$ & PLM \\
% \hline
% \hline
% $S_{GRASP\_noisy}$ $E_0$& 66.7  & {\bf 75.51} \\
% $S_{GRASP\_noisy}$ $E_1$& 66.7  & {\bf 75.51} \\
% $S_{ROBOT}$ $E_0$& 66.7 & 66.7 \\
% $S_{ROBOT}$ $E_1$& 66.7 & 66.7 \\
% 
% \hline
% \end{tabular}
% \caption{Pipeline accuracy $acc_{p}$ (\%): Pre-grasp prediction.}
% \label{tab:pipelineAccuracyGraspHyp}
% \end{table}

Finally, we present results for the grasp and place action for each level of complexity on the $S_{ROBOT}$ dataset in Table~\ref{tab:realRobotGlobalStats}. It is important to notice the performance drop of 20\% in average when the complexity increases from medium (scenario2) to complex (scenario3). Considering that the multiple object scenario of these experiments is rather simple, one way to improve the performance would be to plan for object displacement (sequence of actions before the pre-grasp pose) before the grasp execution in order to increase the grasping performance on scenario3.

\begin{table}[t]
\centering
\begin{tabular}{|c|c|c|c|c|}
\hline
\multirow{2}{*}{Scenario} & Total tests & Reachable  & Grasped & Placed \\
& & pre-grasps & objects & objects \\
\hline
\hline
%\multirow{3}{*}{$S_{REAL\_noisy}}$   &  gt 50\%  &  63.33 \%& 35.40\% & 42.72\%\\
scenario1   & 10 &  9 (90\%) & 7/9 (77.8\%) &  7/9 (77.8\%)\\
\hline
scenario2 &  15  &  10 (66.7\%)& 7/10 (70\%) & 7/10 (70\%)\\
\hline
scenario3 &  20  &  16 (80\%)& 8/16 (50\%) & 8/16 (50\%)\\
\hline
\end{tabular}
%\caption{(Perturbed) Pipeline vs. local shape information (PFH) for the ORCA scenarios.}
\caption{Percentage of successful grasps in the real robot scenarios. Different levels of $S_{ROBOT}$ complexity. }
\label{tab:realRobotGlobalStats} 
\end{table}

We note that in all our experiments with the pipeline \emph{the result changes if the data is perturbed} for the task, pre-grasp and the grasping point prediction. This highlights the importance of the object categorical information for robot grasping. However, when this information is absent the probabilistic logic theory is robust enough to give reasonable results thanks to the use of world general knowledge. The experiments show the \emph{robustness of the probabilistic logic module when more general rules} are used.

%\subsection{Qualitative results using a real robot}

%\todo{Plinio, please add qualitative results here}
   
%%%%%%%%%%%%%%%%%%%%%%%%%%%%%%%%%%%%%%%%%%%%%%%%%%%%%%%%%%%%%%%%%%%%%%%%%%%%%%%%
\section{Conclusions}

We proposed a new probabilistic logic pipeline which combines high-level reasoning and low-level learning for robot grasping. The high-level reasoning leverages symbolic world knowledge, in the form of object/task ontologies and object-task affordances, object categorical and task-based information. The low-level part is based on learning with shape visual features. The non-trivial realization of high-level knowledge relies on logic, which exploits world knowledge and relations to encode compact grasping models that generalize over similar object parts and object/task categories in a natural way. When combined with probabilistic reasoning, our proposed pipeline shows robustness to the uncertainty in the world and missing information. In addition, our experiments confirm the importance of high-level reasoning and world-knowledge as opposed to using solely local shape information for robot grasping.

As future work we point out three important directions. One is learning the parameters and structure of our logic theory from data. Further, we would like to exploit recent advances in the research on persistent homology \cite{Zhu13}. Persistent homology is a tool from topological data analysis performing multi-scale analysis on  point clouds identifying clusters, holes, and voids therein. Incorporating such a technique into our pipeline can improve the current part detector. Another direction is planning a sequence of actions in order to fulfill the task-dependent pre-grasp poses. Since planning in presence of multiple objects raises complexity and generalization issues, considering relational planners similar to those in \cite{394563} may provide successful plans for pre-grasp tasks.


%%%%%%%%%%%%%%%%%%%%%%%%%%%%%%%%%%%%%%%%%%%%%%%%%%%%%%%%%%%%%%%%%%%%%%%%%%%%%%%%

%%%%%%%%%%%%%%%%%%%%%%%%%%%%%%%%%%%%%%%%%%%%%%%%%%%%%%%%%%%%%%%%%%%%%%%%%%%%%%%%

%%%%%%%%%%%%%%%%%%%%%%%%%%%%%%%%%%%%%%%%%%%%%%%%%%%%%%%%%%%%%%%%%%%%%%%%%%%%%%%%
%\section*{APPENDIX}

%\section*{ACKNOWLEDGMENT}

%%%%%%%%%%%%%%%%%%%%%%%%%%%%%%%%%%%%%%%%%%%%%%%%%%%%%%%%%%%%%%%%%%%%%%%%%%%%%%%%

\bibliography{task_dependent_grasping}
\bibliographystyle{IEEEtran}
\end{document}